\documentclass[12pt,onecolumn,oneside,letterpaper,journal]{IEEEtran}
\usepackage{setspace}
\usepackage{amsmath,amsfonts}
\usepackage{algorithmic}
\usepackage{algorithm}
\usepackage{array}
\usepackage[caption=false,font=normalsize,labelfont=sf,textfont=sf]{subfig}
\usepackage{textcomp}
\usepackage{stfloats}
\usepackage{url}
\usepackage{verbatim}
\usepackage{balance}
\usepackage{graphicx}
\usepackage{epstopdf}
\usepackage{cite}
\usepackage{amssymb}
\usepackage{mathrsfs}
\usepackage{float}

\newenvironment{proof}{undefined{\noindent\it Proof:\quad}}{\hfill $\blacksquare$\par}
\newcommand{\sgn}{\mathrm{sgn}}
\allowdisplaybreaks[4]
\begin{document}
\doublespacing
\title{\huge \onehalfspacing Content Popularity Prediction Based on Quantized Federated Bayesian Learning in Fog Radio Access Networks\vspace{-1em}}

\author{Yunwei Tao, Yanxiang~Jiang,~\IEEEmembership{Senior~Member,~IEEE}, ~Fu-Chun~Zheng,~\IEEEmembership{Senior~Member,~IEEE},~Pengcheng~Zhu,~\IEEEmembership{Member,~IEEE},
~Dusit~Niyato,~\IEEEmembership{Fellow,~IEEE}, and Xiaohu~You,~\IEEEmembership{Fellow,~IEEE}\vspace{-4.8em}
\thanks{{Manuscript received \today. This work was supported in part by the National Key Research and Development Program under Grant 2021YFB2900300, the National Natural Science Foundation of China under grant 61971129, and the Shenzhen Science and Technology Program under Grant KQTD20190929172545139. This paper was
presented in part at the IEEE Globecom 2021, Madrid, Spain, December 2021.}}
\thanks{{Yunwei Tao, Yanxiang Jiang, Pengcheng Zhu and Xiaohu You are with the National Mobile Communications Research Laboratory, Southeast University, Nanjing 210096, China, and Yanxiang Jiang is also with the School of Electronic and Information Engineering, Harbin Institute of Technology, Shenzhen 518055, China (e-mail: tarzan\_yw@seu.edu.cn, yxjiang@seu.edu.cn, p.zhu@seu.edu.cn, xhyu@seu.edu.cn).}}
\thanks{{Fu-Chun Zheng is with the School of Electronic and Information Engineering, Harbin Institute of Technology, Shenzhen 518055, China, and the National Mobile Communications Research Laboratory, Southeast University, Nanjing 210096, China (e-mail: fzheng@ieee.org).}}
\thanks{{Dusit Niyato is with the School of Computer Engineering, Nanyang
Technological University (NTU), Singapore 639798, Singapore (e-mail: dniyato@ntu.edu.sg).}}}


\maketitle
\begin{abstract}
\vspace{-1em}
In this paper, we investigate the content popularity prediction problem in cache-enabled fog radio access networks (F-RANs). In order to predict the content popularity with high accuracy and low complexity, we propose a Gaussian process based regressor to model the content request pattern. Firstly, the relationship between content features and popularity is captured by our proposed model. Then, we utilize Bayesian learning to train the model parameters, which is robust to overfitting. However, Bayesian methods are usually unable to find a closed-form expression of the posterior distribution. To tackle this issue, we apply a stochastic variance reduced gradient Hamiltonian Monte Carlo (SVRG-HMC) method to approximate the posterior distribution. To utilize the computing resources of other fog access points (F-APs) and to reduce the communications overhead, we propose a quantized federated learning (FL) framework combining with Bayesian learning. The quantized federated Bayesian learning framework allows each F-AP to send gradients to the cloud server after quantizing and encoding. It can achieve a tradeoff between prediction accuracy and communications overhead effectively. Simulation results show that the performance of our proposed policy outperforms the existing policies.
\end{abstract}
\vspace{-1.5em}
\begin{IEEEkeywords}
\vspace{-1.5em}
F-RANs, Bayesian learning, federated learning, content popularity, content feature.
\end{IEEEkeywords}

\section{Introduction}
\IEEEPARstart{T}{he} 6th-Generation mobile communications system  aims to provide mobile communications services with high quality of service (QoS) and large coverage areas\cite{Towards}. It is also expected to stimulate new developments in various industry applications, such as healthcare, energy, agriculture, etc. Rapid growth increases the pressure on traffic, and fog radio access network (F-RAN) equipped with cache storage is considered as a promising solution\cite{wangbao,chenxuan,huyabai}. In F-RANs, fog access points (F-APs) with limited caching capacity and computing resources are deployed at the edge of the network, where popular contents can be cached to satisfy most of the users’ requests\cite{confYan}. In view of the caching capacity constraints, the issue of predicting content popularity has attracted more and more attention, as it plays an important role in improving caching efficiency.

Traditional caching update policies, such as first in first out (FIFO), least recently used (LRU) and least frequently used (LFU), are widely used in wired networks with sufficient storage and computing resources\cite{755618,6214173,6736753}, but they are inefficient because they ignore the popularity of content. When considering content popularity, most existing edge caching policies assume that content popularity is already known\cite{wanchaoyi2,wanchaoyi1,cuixiaoting}, which is not realistic, especially when the contents are specific to the domain, e.g., IoT applications. It is therefore important to predict future content popularity based on the information collected by the communications network. In\cite{8531745}, the authors proposed a user preference model to predict content popularity with model parameters learned by online gradient descent (OGD), which was proved to be superior to traditional caching schemes in\cite{755618,6214173,6736753}. In\cite{9354176}, the authors proposed two online prediction models for content popularity, namely the popularity prediction model and the Grassman prediction model, where the unconstrained coefficients for linear prediction were obtained by solving a constrained non-negative least squared method. In\cite{dblp}, the authors investigated the effect of time on the prediction of popularity by dynamically weighting the feature sequences, mainly using dynamic weighting techniques to model the changes in the importance of the feature sequences over time. A simplified bidrectional long short-term memory (Bi-LSTM) network based content popularity prediction scheme was proposed in\cite{9200588}, which tracked popularity trend by the number of requests. In\cite{wuyuting}, federated learning (FL) was utilized to obtain the context-aware popularity prediction model. In\cite{8885590}, popular contents were learned through a Possion regressor model whereby Monte Carlo method was applied to optimize the model parameters. However, in F-RANs, it is difficult to obtain adequate data about numerous contents through F-APs with limited storage. These existing works except \cite{wuyuting} and \cite{8885590} fail to predict the popularity of newly-added contents with few statistical data. In other words, the content popularity cannot be predicted with a small number of training samples\cite{8917719}. For example, the performance was limited by lack of statistical data\cite{overfitting}, which caused serious overfitting problems. Moreover, most of the prediction policies employ centralized learning approach that cannot efficiently utilize the computing resources of edge devices.


Most current research works focus on prediction over a single F-AP or high power node (HPN). However, the coverage area and statistical data of a single F-AP are limited, so multiple F-APs need to collaborate in prediction\cite{wuyuting}. However, collaborative prediction models suffer from a critical issue. In\cite{9599369,9415623}, the authors claimed that transferring data from data owners to the server would lead to privacy leakage, and it was also impractical to transfer all data to the centralized server due to limited network bandwidth. Correspondingly, FL, a novel distributed learning approach that does not need to exchange raw user data, was proposed to solve these problems\cite{wuyuting}. FL enables multiple participating devices to jointly train a machine learning model. Each participant computes the local model based on the local training data and sends the local model to the server. Nowadays, FL has become an effective framework to perform distributed optimization.

Despite the benefits brought by FL, communications is still the challenge to confront with. In order to deal with the communications overhead in FL training process, much research about communications in FL has been conducted. In\cite{8917724}, the authors proposed an adaptive federated averaging method using distributed Adam optimization and compression technique. In\cite{8889996}, a sparse ternary compression method was proposed to meet the requirements of FL, which extended the existing compression of Top-$K$ gradient sparsification with optimal Golomb encoding. In\cite{CommunicationEfficient}, a quantization method was proposed to reduce the volume of the model parameters exchanged among devices, and an efficient wireless resource allocation scheme was developed. In\cite{abs190913014}, the authors used periodic averaging, partial device participation and quantized message passing to achieve a tradeoff between communications and computation. In conclusion, communications efficiency is an important problem that has to be considered in FL.

Motivated by the aforementioned discussions, we propose a popularity prediction policy via content request probabilistic model based on content features. The model parameters are learned by Bayesian learning, which is a robust method against overfitting caused by lack of statistical data\cite{9165221}. The training inputs of the prediction model are the features of contents requested by users and the number of requests in different time periods. Due to the computing resources constraint in F-APs, a modified stochastic gradient based Bayesian learning method is adopted to learn the popularity prediction model. Considering the collaboration of F-APs, we combine the advantages of Bayesian learning and FL to establish a distributed learning framework. Besides, we propose a quantization and encoding scheme to improve the communications efficiency in FL. 
Our main contributions are summarized as follows.
\begin{itemize}
	\item[1)]We build a content request model and propose a content popularity prediction policy based on content features and the number of content requests. The proposed policy does not require user preferences and therefore needs less access to user privacy. Then, the content request model is trained by Bayesian learning to obtain the predicted popularity. Specifically, we innovatively introduce stochastic variance reduced gradient (SVRG) into the traditional Bayesian learning method, which can reduce the computational complexity and training time effectively. Finally, our proposed popularity prediction policy can predict the future content popularity of existing contents as well as newly-added contents.
	\item[2)]We propose a global content popularity prediction model based on an FL framework. The method integrates Bayesian learning into the FL framework. In this way, the local models can be generated by Bayesian learning and finally aggregated effectively to obtain the global model, which greatly reduces the computational burden and communications overhead of the cloud server. Besides, our proposed model is robust to overfitting due to the local Bayesian learning procedure. 
	\item[3)]We propose a quantization and encoding scheme for federated Bayesian learning to reduce communications overhead. The gradients uploaded by participants are quantized. After that, a variable-length encoding method is used to encode the quantization results for transmission with improved communications efficiency. We analyze the performance of the quantization and encoding scheme and the upper bound on the number of bits required for transmission after encoding. The theoretical analysis shows that our proposed scheme can further reduce the communications overhead and achieve an efficient tradeoff between accuracy and communications overhead.
	\item[4)]We validate our theoretical results with real dataset. The simulation results show that our proposed quantized federated Bayesian learning based content popularity prediction policy can predict the future content popularity with high accuracy and refresh the cache storage based on the prediction results. Compared with traditional policies, a higher cache hit rate and lower communications overhead are achieved.
\end{itemize}

The reminder of this paper is organized as follows. The system model is presented in Section II. Section III describes the proposed popularity prediction policy based on quantized federated Bayesian learning. Simulation results are given in Section IV and final conclusions are drawn in Section V.

\section{System Model}
As shown in Fig. 1, we consider an F-RAN architecture with $M$ F-APs, where the users are served by these F-APs and the cloud server. Let $\mathcal{Q}=\left\{ q_1,q_2,...,q_m,...,q_M \right\} $ denote the set of F-APs and $\mathcal{C}=\left\{ c_1,c_2,\dots ,c_f,\dots ,\left. c_F \right\} \right. $ the content library located in the cloud server. Let $\boldsymbol{x}_f$ denote the feature vector of content $c_f$ and $s_f$ the size of content $c_f$. Take the movie category as an example, and a file may include genre (e.g., action, comedy, and sci-fi) and some other features such as release year and production company. If content $c_f$ requested by a user is stored in its associated F-AP, then a cache hit occurs, and the content can be downloaded directly from that F-AP. Otherwise, the content needs to be fetched from the neighboring F-APs or the cloud server.
\begin{figure}[b!]
\vspace{-1.5em}
\centering
\includegraphics[width=0.6\textwidth,height=0.25\textwidth]{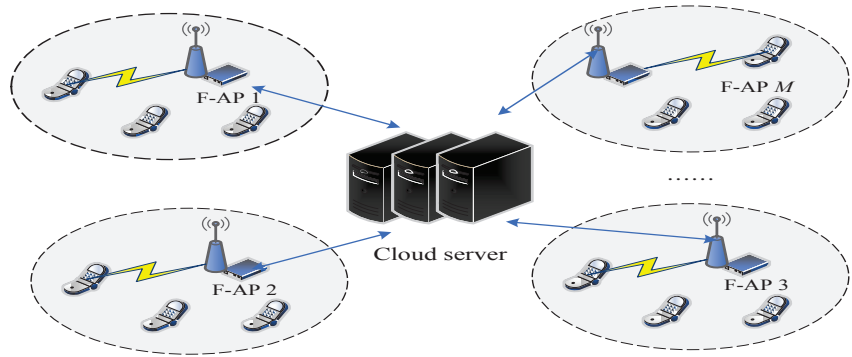}
\caption{Illustration of the scenario in F-RAN.}
\end{figure}

F-AP can collect information about the user's requested contents, which mainly includes the features of the requested contents and the number of requests. We define $\boldsymbol{r}_c\left[ n \right] =\left[ r_{c_1}\left[ n \right] ,\dots ,r_{c_F}\left[ n \right] \right] ^T$ as a vector representing the number of requests for each content in the content library during the $n$-th time period. Considering that the number of requests shows the level of preference for movies, we take the number of requests as the popularity indicator for the contents. Assume that $N$ time periods have been observed. Let $r_{c_f}^{*}\left[ N+1 \right] $ and $r_{c_f}\left[ N+1 \right] $ denote the predicted and real numbers of requests of $c_f$ during the ($N+1$)-th time period, respectively. There exists deviation between the predicted and real numbers of requests. Correspondingly, the root mean squared error (RMSE) is utilized to measure the accuracy of the prediction as follows:
\begin{equation}\label{1}
\setlength{\abovedisplayskip}{3pt}
\setlength{\belowdisplayskip}{3pt}
\mathrm{RMSE}=\sqrt{\frac{1}{F}\sum_{f=1}^F{\left| r_{c_f}^{*}\left[ N+1 \right] -r_{c_f}\left[ N+1 \right] \right|^2}}.
\end{equation}

The cache hit rate is defined as the ratio of the number of cache hits to the number of total requests. The higher the number of cache hits is, the less contents are fetched from the cloud server, which results in less communications pressure over the backhaul links. By caching contents with higher popularity, the cache hit rate can be significantly increased.


The goal of this paper is to find a content popularity prediction policy that minimizes the prediction RMSE with low computational complexity and reduces communications overhead.

\section{Proposed Content Popularity Prediction Policy Based on Quantized Federated Bayesian Learning}
In order to minimize the prediction RMSE, we propose a quantized federated Bayesian learning based content popularity prediction policy in this section, which is composed of a probabilistic model construction phase, a local model learning phase, a global model integration phase, and a predicting phase. The proposed policy utilizes the number of requests and content features, and can predict content popularity accurately with low computational complexity and small communications overhead.
\subsection{Policy Outline}
\subsubsection{Probabilistic model construction phase}
The relationship between content popularity and content features is hard to generalize. Given the nonlinear relationship between content features and content popularity, traditional regression models are not able to capture it. Gaussian process, on the other hand, can model nonlinear relationship flexibly and effectively. Therefore, the construction of a probabilistic regression model based on Gaussian process is the first step in our proposed policy. If the feature vectors of contents are similar, the prevalence of the two contents will be close to each other. The request arrivals are modeled using the typical Poisson distribution, which is the most commonly used and intuitive model.
\subsubsection{Local model learning phase}
In the edge caching structure in F-RANs, F-APs may obtain only a few request observations so overfitting is a challenging problem\cite{9165221}. We propose to learn the probabilistic model via Bayesian learning due to its robustness to overfitting\cite{8917719}. However, Bayesian learning cannot yield a closed-form expression in most cases. Therefore, Markov chain Monte Carlo (MCMC) is utilized to approximate the model parameters. Concretely, we apply an SVRG based variant of MCMC to content popularity prediction for the first time. First, we obtain the unnormalized posterior approximation of the parameters based on their prior knowledge. Second, we compute the gradient of the posterior distribution on the basis of the existing request observations accordingly. Finally, we apply a discretization method to approximate the updated parameters via multiple sampling.
\subsubsection{Global model integration phase}
The coverage area and training data of a single F-AP are limited. Consequently, we propose a training approach that combines FL and Bayesian learning. First, each F-AP still takes the Bayesian approach proposed in the second step for local model learning. Second, each F-AP is assigned different weights according to the local dataset size and participates in FL to learn the global model. Finally, a quantization and encoding scheme is proposed for federated Bayesian learning to trade off the communications overhead and prediction accuracy.
\subsubsection{Predicting phase}
After the training and model integration phases, the approximate model parameters can be estimated. Accordingly, F-AP is capable of predicting the popularity of existing contents or newly-added contents in the next time period with low computational complexity.

\subsection{Probabilistic model construction phase}
\subsubsection{Brief introduction to Gaussian process}
Gaussian process is a very powerful nonparametric Bayesian tool for modeling nonlinear problems. A Gaussian process is a set of random variables of a certain class, and any number of random variables in the set have a joint Gaussian distribution\cite{8664622}. Using a Gaussian process, we can define the distribution with a nonparametric function $f\left( \boldsymbol{x} \right) \sim \mathcal{G} \mathcal{P} \left( \mu \left( \boldsymbol{x} \right) ,K\left( \boldsymbol{x},\boldsymbol{x}' \right) \right)$,  
where $\boldsymbol{x}$ is a $Q$-dimensional random variable, $\mu \left( \boldsymbol{x} \right) $ and $K\left( \boldsymbol{x},\boldsymbol{x}' \right) $ are the mean and kernel functions of the Gaussian process, respectively, which are defined as follows:
\begin{equation}\label{5}
\setlength{\abovedisplayskip}{3pt}
\setlength{\belowdisplayskip}{3pt}
\mu \left( \boldsymbol{x} \right) =\mathbb{E}\left[ f\left( \boldsymbol{x} \right) \right], \quad K\left( \boldsymbol{x},\boldsymbol{x}' \right) =\mathbb{E}\left[ \left( f\left( \boldsymbol{x} \right) -\mu \left( \boldsymbol{x} \right) \right) \left( f\left( \boldsymbol{x}' \right) -\mu \left( \boldsymbol{x}' \right) \right) \right].
\end{equation}
This means that any $M$ samples follow a joint Gaussian distribution $\left[ f\left( \boldsymbol{x}_1 \right) ,\cdots ,f\left( \boldsymbol{x}_M \right) \right] ^T\sim \mathcal{N} \left( \boldsymbol{\mu },\boldsymbol{K} \right)$,
where $\boldsymbol{\mu }=\left[ \mu \left( \boldsymbol{x}_1 \right) ,\cdots ,\mu \left( \boldsymbol{x}_M \right) \right] ^T$ denotes the mean vector and $\boldsymbol{K}$ represents the covariance matrix $\left[ \boldsymbol{K} \right] _{i,j}=K\left( \boldsymbol{x}_i,\boldsymbol{x}_j \right) $. The kernel function $K$ specifies the features of the function that we wish to model. It depends on intuition and experience to choose a proper kernel function for a learning task, and the squared exponential kernel functions, Marton kernel functions and radial basis kernel functions are most commonly used.

Gaussian process is often used in prediction tasks. By giving a set of training datasets $\mathcal{D} \triangleq \left\{ \boldsymbol{X},\boldsymbol{Y} \right\} $, where $\boldsymbol{Y}=\left[ \boldsymbol{y}_1,...,\boldsymbol{y}_n \right] ^T$ is the training output and $\boldsymbol{X}=\left[ \boldsymbol{x}_1,...,\boldsymbol{x}_n \right] $ is the training input, the aim is to predict the output $\boldsymbol{Y}^{\ast}=\left[ \boldsymbol{y}_{1}^{\ast},...,\boldsymbol{y}_{n^{\ast}}^{\ast} \right] ^T$ given a test input dataset $\boldsymbol{X}^{\ast}=\left[ \boldsymbol{x}_{1}^{\ast},...,\boldsymbol{x}_{n^{\ast}}^{\ast} \right] $ and the posterior distribution $p\left( \boldsymbol{Y}^{\ast}|\mathcal{D} ,\boldsymbol{X}^{\ast};\boldsymbol{\theta } \right) $, where $\boldsymbol{\theta}$ is the hyper-parameter vector in the kernel function. According to the definition of Gaussian process, the training output $\boldsymbol{Y}$ and the test output $\boldsymbol{Y}^{\ast}$ follow the joint distribution:
\begin{equation}\label{7}
\setlength{\abovedisplayskip}{3pt}
\setlength{\belowdisplayskip}{3pt}
\left[ \begin{array}{c}	\boldsymbol{Y}\\	\boldsymbol{Y}^{\ast}\\\end{array} \right] \sim \mathcal{N} \,\,\left( \mathbf{0},\left[ \begin{matrix}	\boldsymbol{K}+\sigma _{e}^{2}\boldsymbol{I}_n&		\boldsymbol{k}^{\ast}\\	(\boldsymbol{k}^{\ast})^{T}&		\boldsymbol{k}^{\ast \ast}\\\end{matrix} \right] \right),
\end{equation}
where $\sigma _e$ is the hyper-parameter of the Gaussian process, $\boldsymbol{I}_n$ represents the $n$-dimensional identity matrix, $\boldsymbol{K}=K\left( \boldsymbol{X},\boldsymbol{X} \right) $ represents the $n\times n$ correlation matrix between the training inputs, $\boldsymbol{k}^{\ast}=K\left( \boldsymbol{X},\boldsymbol{X}^{\ast} \right) $ represents the $n\times n^{\ast}$ correlation matrix between the training and test inputs, and $\boldsymbol{k}^{\ast \ast}$ represents the $n^{\ast}\times n^{\ast}$ correlation matrix between the test inputs. By applying the properties of the Gaussian distribution, the variance and mean of the posterior distribution can be obtained as follows:
\begin{equation}\label{8}
\setlength{\abovedisplayskip}{3pt}
\setlength{\belowdisplayskip}{3pt}
\mathbb{E}\left[ \boldsymbol{Y}^{\ast} \right] =(\boldsymbol{k}^{\ast})^{T}\left( \boldsymbol{K}+\sigma _{e}^{2}\boldsymbol{I}_n \right) ^{-1}\boldsymbol{Y},\quad \mathrm{Var}\left[ \boldsymbol{Y}^{\ast} \right] =\boldsymbol{k}^{\ast \ast}-(\boldsymbol{k}^{\ast})^{T}\left( \boldsymbol{K}+\sigma _{e}^{2}\boldsymbol{I}_n \right) ^{-1}\boldsymbol{k}^{\ast}.
\end{equation}
Accordingly, the statistical properties of the predicted output can be obtained.

\subsubsection{Probabilistic model}
In this subsection, we propose a probabilistic model to show the users' request pattern. In order to introduce the proposed model, first, we assume that each content has a set of characteristics. For example, a video may belong to a specific category (e.g., education, arts, entertainment, and technology), and have some other features such as year of release, language, etc. Let $\boldsymbol{x}_f$ denote a $Q$-dimensional feature vector of content $c_f$ whose values can be binary or continuous, which allows the proposed probabilistic model to handle diverse combinations of features. Then, the proposed regression-based hierarchical multilevel probabilistic model can be expressed as follows:
\begin{subequations}\label{2}
\setlength{\abovedisplayskip}{3pt}
\setlength{\belowdisplayskip}{3pt}
\begin{flalign}
\label{modela}&\boldsymbol{r}_{c_f}\left[ n \right] \left| \lambda _f\left( \boldsymbol{x}_f \right) \right. \sim \mathrm{Poisson}\left( e^{\lambda _f\left( \boldsymbol{x}_f \right)} \right) ,\forall n=1,\dots ,N,\\
\label{modelb}&\lambda _f\left( \boldsymbol{x}_f \right) \,\,\left| g\left( \boldsymbol{x}_f \right) ,\beta _0 \right. \sim \mathcal{N} \left( g\left( \boldsymbol{x}_f \right) ,\beta _0 \right) ,\\
\label{modelc}&g\left( \boldsymbol{x} \right) \left| \boldsymbol{x},\beta _1,\dots , \right. \beta _{Q+1}\sim \mathcal{G} \mathcal{P} \left( 0,K\left( \boldsymbol{x},\boldsymbol{x}' \right) \right) ,\\
\label{modeld}&\beta _q\left| \mathrm{Gamma}\left( A_q,B_q \right) \right. ,\forall q=0,\dots ,Q+1.
\end{flalign}
\end{subequations}

The first layer of the model is the distribution of content request observations. In this layer, requests for content $c_f$ are assumed to obey a Poisson distribution (\ref{modela}) with natural parameter $\lambda _f$, and the request rate $e^{\lambda _f \left( \boldsymbol{x}_f \right)}$ is an exponential function of the natural parameters. The Poisson distribution is a simple and widely used distribution to describe the occurrence number of a random event during unit time (or space) for modeling count data.

As mentioned in the previous discussions, the model recognizes that those contents with similar features tend to have similar request patterns or popularity, and this prior information needs to be applied at a higher level of the model. In (\ref{modelb}), $\lambda _f\left( \boldsymbol{x}_f \right) $ follows a normal distribution with mean $g\left( \boldsymbol{x}_f \right) $ and variance $\beta _0$\cite{uncertain}. With this assumption, the model defines popularity as a function of content features and also allows contents with the same features to have different popularity  through the variance $\beta _0$, which is consistent with prior knowledge.

In the third layer (\ref{modelc}) of the model, the Gaussian process is applied and $g\left( \boldsymbol{x}_f \right) $ is assumed to be the realization of a Gaussian process with the zero mean and a covariance matrix with kernel function $K\left( \boldsymbol{x}_i,\boldsymbol{x}_j \right) $, which captures the relationship between the different features and reflects them on the subsequent samples.  The covariance matrix $\left[ \boldsymbol{K} \right] _{i,j}=K\left( \boldsymbol{x}_i,\boldsymbol{x}_j \right) $ is the kernel function that determines the interpretation of the Gaussian process and determines the correlation or similarity between the random variables. Our proposed model utilizes the squared exponential kernel (SEK), which is defined as $K\left( \boldsymbol{x}_i,\boldsymbol{x}_j \right) =\beta _1e^{-\sum_{q=2}^{Q+1}{\beta _q\left\| x_{i}^{q-1}-x_{j}^{q-1} \right\| ^2}}$,
where $\beta _1$ determines the vertical scale and $\beta _q$ determines the horizontal scale on the $q$-th dimensional feature. By using a different scale for each input feature dimension, the model can change their importance. For example, if $\beta _q$ tends to 0, the features in the $q$-th dimension have little effect on the covariance. Besides, the kernel function $K\left( \boldsymbol{x}_i,\boldsymbol{x}_j \right)$ is infinitely differentiable due to the nature of the exponential function.

Finally, the model introduces uncertainty in the values of the kernel function parameters $\beta _1\sim \beta _{Q+1}$ and the variance of the Gaussian distribution $\beta _0$, since existing prior knowledge cannot completely determine their values. The gamma distribution $\mathrm{Gamma}\left( A_q,B_q \right) $ is used as the prior for each of them, where $A_q$ is called the shape parameter and $B_q$ is the inverse scale parameter.

\begin{figure}[b!]
\vspace{-1.5em}
\centering
\includegraphics[scale=0.4]{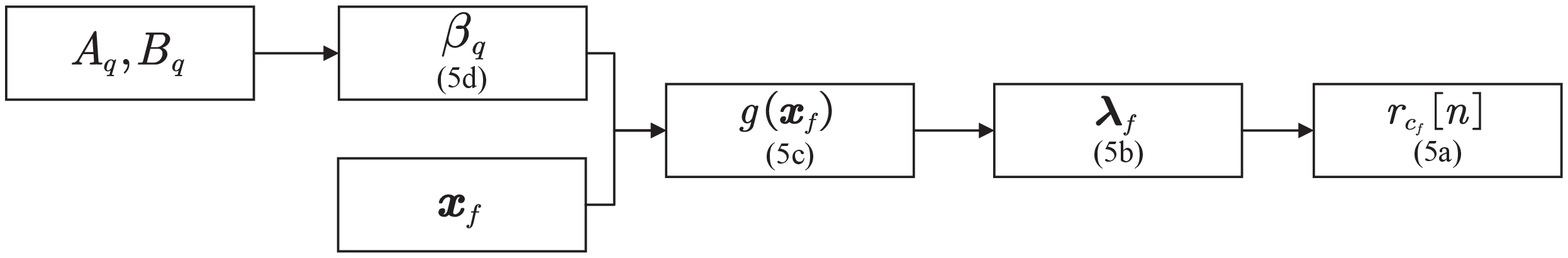}
\caption{The schematic of our proposed probabilistic model.}
\label{mapping}
\end{figure}

The schematic of our proposed probabilistic model is shown in Fig. \ref{mapping}, which indicates the mapping relationship between the parameters at each level of the probabilistic model. In our proposed probabilistic model, the number of requests for each content is assumed to obey a Poisson distribution in (\ref{modela}), and its request rate in (\ref{modelb}) is determined by a realization of Gaussian process in (\ref{modelc}). The realization is related to content features and kernel function parameters of the Gaussian process in (\ref{modeld}). Finally, the model uncertainty is measured by the Gamma distribution with $A_q$ and $B_q$.

\subsection{Local model learning phase}
\subsubsection{Bayesian learning in a nutshell}
Bayesian learning is an important branch in machine learning, which is based on Bayesian theory. The Bayesian theory states that the model parameters follow a potential distribution, and we can make optimal decisions of parameters based on this distribution and the observed data. The most important advantage is that it integrates the prior knowledge of model parameters, which makes the model more robust to overfitting.
Given a set of training examples $\mathcal{D} $, we need to infer the model $\boldsymbol{h}$ that generates these data. The model $\boldsymbol{h}$ is considered to be jointly determined by the objective function and the parameters $\boldsymbol{\omega }$ of the distribution. Since the objective function is usually specified before training, we are thus more interested in $\boldsymbol{\omega }$. In other words, we can infer the model $\boldsymbol{h}$ by estimating the parameters $\boldsymbol{\omega }$, i.e., $\boldsymbol{h}=\mathrm{arg}\max \left\{ P\left( \boldsymbol{\omega }\left| \mathcal{D} \right. \right) ,\boldsymbol{h}\in \mathcal{H} \right\} $, where $\mathcal{H} $ is the set of all possible models.

The model parameters that maximize the posterior distribution are chosen as the best estimate values, and the Bayesian rule provides an effective way to calculate the maximum probability as follows:
\begin{equation}\label{9}
\setlength{\abovedisplayskip}{3pt}
\setlength{\belowdisplayskip}{3pt}
P\left( \boldsymbol{\omega }\left| \mathcal{D} \right. \right) =\frac{P\left( \boldsymbol{\omega } \right) P\left( \mathcal{D} \left| \boldsymbol{\omega } \right. \right)}{P\left( \mathcal{D} \right)}=\frac{P\left( \boldsymbol{\omega } \right) P\left( \mathcal{D} \left| \boldsymbol{\omega } \right. \right)}{\int_{\boldsymbol{\omega }}{P\left( \boldsymbol{\omega } \right) P\left( \mathcal{D} \left| \boldsymbol{\omega } \right. \right) \mathrm{d}\boldsymbol{\omega }}},
\end{equation}
where $P\left( \boldsymbol{\omega } \right) $ is the prior distribution, $P\left( \boldsymbol{\omega }\left| \mathcal{D} \right. \right) $ is the posterior distribution based on training data, and $P\left( \mathcal{D} \right) $ is the normalized constant.

\subsubsection{Model learning}
Bayesian learning based approach is proposed to train the probabilistic model. The aim is to update the estimates of the model parameters $\left\{ \lambda _f\left( \boldsymbol{x}_f \right) \right\} _{f=1}^{F}$ , $g\left( \boldsymbol{x} \right) $ and $\left\{ \beta _q \right\} _{q=0}^{Q+1}$ given the content request observation set $\mathcal{R} =\left\{ \boldsymbol{r}_c\left[ n \right] \right\} _{n=1}^{N}$. However, since $g\left( \boldsymbol{x} \right) $ is an infinite-dimensional vector, it is difficult to estimate. Therefore, we instead focus on its realizations $\left\{ g\left( \boldsymbol{x}_f \right) \right\} _{f=1}^{F}$ . To simplify the reasoning, the following distribution can be formulated from the proposed model:
\begin{equation}\label{10}
\setlength{\abovedisplayskip}{3pt}
\setlength{\belowdisplayskip}{3pt}
\left[ \lambda _1\left( \boldsymbol{x}_1 \right) ,\dots ,\lambda _f\left( \boldsymbol{x}_f \right) ,\dots ,\lambda _F\left( \boldsymbol{x}_F \right) \right] ^T\sim \mathcal{N} \left( 0,\boldsymbol{K}' \right),
\end{equation}
where $\boldsymbol{K}'=\boldsymbol{K}+\beta _0\boldsymbol{I}$. By combining Bayesian rule and the model assumptions, the posterior distribution of the unknown parameters of the proposed model can be obtained as follows:
\begin{equation}\label{11}
\setlength{\abovedisplayskip}{3pt}
\setlength{\belowdisplayskip}{3pt}
p\left( \boldsymbol{\lambda },\boldsymbol{\beta }\left| \mathcal{R} \right. \right) =p\left( \mathcal{R} \left| \boldsymbol{\lambda } \right. \right) p\left( \boldsymbol{\lambda }\left| \boldsymbol{\beta } \right. \right) \prod_{q=0}^{Q+1}{p\left( \beta _q \right)}/H,
\end{equation}
where $\boldsymbol{\beta }$ and $\boldsymbol{\lambda }$ represent the proposed model parameters $\boldsymbol{\lambda }=\left[ \lambda _1\left( \boldsymbol{x}_1 \right) ,\dots ,\lambda _f\left( \boldsymbol{x}_f \right) ,\dots ,\lambda _F\left( \boldsymbol{x}_F \right) \right] ^T$ and $\boldsymbol{\beta }=\left[ \beta _0,\dots ,\beta _q,\dots \beta _{Q+1} \right] ^T$, respectively, $p\left( \mathcal{R} |\boldsymbol{\lambda } \right) =\prod_{n=1}^N{\prod_{f=1}^F{p\left( r_{c_f}\left[ n \right] |\lambda _f \right)}}$, $p\left( \boldsymbol{\lambda },\boldsymbol{\beta }\left| \mathcal{R} \right. \right) $ is the posterior distribution based on historical request observations and prior knowledge, and $H$ is the normalized constant. It usually requires very complex integration operations of the marginal likelihood function to obtain the normalized constant $H$. Therefore, it is impractical to obtain a closed expression for the posterior distribution, meaning that the optimal parameters cannot be obtained analytically. To solve this problem, an efficient sampling method is required that samples from the unnormalized posterior distribution and then calculates over multiple random samples to approximate the true posterior distribution and estimate the parameters. 

For the above task, MCMC is an effective tool for training our proposed probabilistic model. The goal of MCMC  is to generate a set of independent samples from the unnormalized posterior distribution, which contains enough samples to perform inference. However, the traditional MCMC method has some limitations, such as it needing long time to reach the stationary distribution. Therefore, Hamiltonian Monte Carlo (HMC) and its variants have been  chosen for sampling\cite{hmc,8885590}, which has been one of the most successful methods in MCMC for sampling from unnormalized distributions.

HMC is a popular method based on the simulation of Hamiltonian dynamics and can generate a sequence of samples $\left\{ \boldsymbol{\xi }_s \right\} _{s=1}^{S}$ from a $U$-dimensional distribution $p\left( \boldsymbol{\xi } \right) $. In order to generate the sample $\boldsymbol{\xi }_{s+1}$, the sample $\boldsymbol{\xi }_s$ is considered as the position of a particle and their directions of motion are determined by the Hamiltonian dynamics. The position of $\boldsymbol{\xi }_{s+1}$ is obtained by simulating the motion of the particle. HMC uses Hamiltonian dynamics to construct Markov chains and introduces a $U$-dimensional auxiliary momentum variable $\boldsymbol{\theta }$. HMC is only applicable to differentiable functions and real-valued unconstrained variables. However, $\beta _q$ in the kernel function must be positive so that the exponential transformation $\rho _q=\log \left( \beta _q \right) $ can be used to make $\rho _q$ an unconstrained auxiliary variable. Let $\boldsymbol{\tau }=\left[ \boldsymbol{\lambda }^T,\rho _0,...,\rho _{Q+1} \right] ^T\in R^{\left( F+Q+2 \right)}$ denote the vector of unknown parameters and $\boldsymbol{r}_{c_f}=\left[ r_{c_f}\left[ 1 \right] ,r_{c_f}\left[ 2 \right] ,\cdots ,r_{c_f}\left[ N \right] \right] ^T$ the request observation vector for content $c_f$ during $N$ time periods. Accordingly, by substituting the prior knowledge and the established model, the negative logarithm of the posterior distribution can be derived from Eq. (\ref{11}) as follows:
\begin{align}\label{12}
\setlength{\abovedisplayskip}{3pt}
\setlength{\belowdisplayskip}{3pt}
\phi \left( \boldsymbol{\tau } \right) &=-\log p\left( \boldsymbol{\lambda },\boldsymbol{\beta }|\mathcal{R} \right)=-\log p\left( \mathcal{R} |\boldsymbol{\lambda } \right) -\log p\left( \boldsymbol{\lambda }|\boldsymbol{\beta } \right) -\sum_{q=0}^{Q+1}{\log p\left( \beta _q \right)}+\log H\\
&=\sum_{n=1}^N{\sum_{f=1}^F{-r_{c_f}\left[ n \right] \lambda _f+e^{\lambda _f}}}+\frac{1}{2}\log\det \left( \boldsymbol{K}' \right)+\frac{1}{2}\boldsymbol{\lambda }^T\boldsymbol{K}'^{-1}\boldsymbol{\lambda }
+\sum_{q=0}^{Q+1}{-A_q\rho _q+B_qe^{\rho _q}}+\log H.\nonumber
\end{align}
The gradients of Eq. (\ref{12}) are required to perform HMC sampling, which can be derived as follows:
\begin{align}\label{13}
\setlength{\abovedisplayskip}{3pt}
\setlength{\belowdisplayskip}{3pt}
\nabla \phi \left( \boldsymbol{\tau } \right) &=-\sum_{n=1}^N{\sum_{f=1}^F{\nabla \log p\left( r_{c_f}\left[ n \right] |\lambda _f \right)}}-\nabla \log p\left( \boldsymbol{\lambda }|\boldsymbol{\beta } \right)-\sum_{q=0}^{Q+1}{\nabla \log p\left( \beta _q \right)},\notag\\
\frac{\partial \phi \left( \boldsymbol{\tau } \right)}{\partial \lambda _f}&=\sum_{n=1}^N{-r_{c_f}\left[ n \right] +Ne^{\lambda _f}}+\left[ \boldsymbol{K}'^{-1}\boldsymbol{\lambda } \right] _f,\notag\\
\frac{\partial \phi \left( \boldsymbol{\tau } \right)}{\partial \rho _q}&=\frac{1}{2}\mathrm{tr}\left( \boldsymbol{K}'^{-1}\frac{\partial \boldsymbol{K}'}{\partial \rho _q} \right) -\frac{1}{2}\boldsymbol{\lambda }^{\mathrm{T}}\boldsymbol{K}'^{-1}\frac{\partial \boldsymbol{K}'}{\partial \rho _q}\boldsymbol{K}'^{-1}\boldsymbol{\lambda }-A_q+B_qe^{\rho _q}.
\end{align}

It can be observed that a notable drawback of HMC is that it must traverse the entire dataset for gradient computation. However, it is hard for F-APs to handle huge computational tasks due to the limited computing resources. To solve this problem, we consider the gradient of the posterior distribution. The first term in Eq. (\ref{13}) requires traversing the entire dataset to find the gradient, which is the main reason for the large computational effort of HMC. Therefore, we propose an improved method based on stochastic gradient to randomly select subset $\tilde{\mathcal{R}}$ from the dataset $\mathcal{R}$:
\begin{align}\label{15}
\setlength{\abovedisplayskip}{3pt}
\setlength{\belowdisplayskip}{3pt}
\nabla \tilde{\phi}\left( \boldsymbol{\tau } \right) &=-\frac{\left| \mathcal{R} \right|}{\left| \tilde{\mathcal{R}} \right|}\sum_{\boldsymbol{r}_c\in \mathcal{R}}^{\,\,}{\nabla \log p\left( r_{c_f}\left[ n \right] |\lambda _f \right)}-\nabla \log p\left( \boldsymbol{\lambda }|\boldsymbol{\beta } \right)-\sum_{q=0}^{Q+1}{\nabla \log p\left( \beta _q \right)}.
\end{align}

The stochastic gradient introduces noise to the proposed model. According to the central limit theorem, the noisy gradient can be approximated as $\nabla \tilde{\phi}\left( \boldsymbol{\tau } \right) \approx \nabla \phi \left( \boldsymbol{\tau } \right) +\mathcal{N} \left( 0,V\left( \boldsymbol{\tau } \right) \right)$\cite{CTQ},
where $V\left( \boldsymbol{\tau } \right) $ is the covariance matrix of the noise introduced by the stochastic gradient, which depends on the dimension of model parameters and sample size. Simply using stochastic gradient will lead to a large variance, which can reduce the convergence speed of HMC algorithm. The SVRG method can effectively reduce the impact of the variance and thus shorten the model training time. The SVRG-HMC based sample update can be expressed as follows\cite{MyConf}:
\begin{equation}\label{17}
\setlength{\abovedisplayskip}{3pt}
\setlength{\belowdisplayskip}{3pt}
\boldsymbol{\theta }_{s+1}=\left( 1-Dh \right) \boldsymbol{\theta }_s-h\tilde{\nabla}_t+\sqrt{2Dh}\cdot \boldsymbol{\eta }_t, \quad\boldsymbol{\xi }_{s+1}=\boldsymbol{\xi }_s+h\boldsymbol{\theta }_{s+1}.
\end{equation}

\begin{algorithm}[t!]
\algsetup{linenosize=\scriptsize} \footnotesize
\caption{ The proposed SVRG-HMC sampling method. }
\begin{algorithmic}[1]
\REQUIRE $S, L, b, h>0, Dh<1, D\geqslant 1$;
\ENSURE $\left\{ \boldsymbol{\tau }_s \right\} _{1\leqslant s\leqslant S/L-1}$;
\STATE Initialize $\boldsymbol{\theta }_0, \boldsymbol{\xi }_0$;
\FOR {$s=0,1,...,S/L-1$}
\STATE $g=-\frac{1}{NF}\sum\nolimits_{i=1}^{NF}{\nabla \log p\left( r_{c_f}\left[ n \right] |\lambda _f \right)}$
\STATE $  \,\,\,\,\,=\frac{1}{NF}\sum\nolimits_{i=1}^{NF}{\nabla \delta _i\left( \boldsymbol{\xi } \right)}$;
\STATE $\boldsymbol{\omega }=\boldsymbol{\xi }_{sL}$;
\FOR {$l=0,1,...,L-1$}
\STATE Randomly draw a subset $I\subseteq \mathcal{R} , \left| I \right|=b$ with equal probability;
\STATE $\tilde{\nabla}_{sL+l}=-\log p\left( \boldsymbol{\xi }_{sL+l} \right) -\sum_{q=0}^{Q+1}{\log p\left( \boldsymbol{\xi }_{sL+l} \right)}+\frac{NF}{b}\sum_{i\in I}^{\,\,}{\left( \nabla \delta _i\left( \boldsymbol{\xi }_{sL+l} \right) -\nabla \delta _i\left( \boldsymbol{w} \right) \right)}+g$;
\STATE $\boldsymbol{\theta }_{sL+l+1}=\left( 1-Dh \right) \boldsymbol{\theta }_{sL+l}-h\tilde{\nabla}_{sL+l}+\sqrt{2Dh}\cdot \boldsymbol{\eta }_{sL+l}$;
\STATE $\boldsymbol{\xi }_{sL+l+1}=\boldsymbol{\xi }_{sL+l}+h\boldsymbol{\theta }_{sL+l+1}$;
\ENDFOR
\STATE $\boldsymbol{\tau }_s=\boldsymbol{\xi }_{sL+L}$;
\ENDFOR
\end{algorithmic}
\end{algorithm}

The procedure of SVRG-HMC is outlined in Algorithm 1, where $S$ is the sampling round, $b$ the size of minibatch, $L$ the number of discretization steps, $h$ the learning rate and $D$ a constant. Specifically, the algorithm can be divided into inner and outer loops, where the outer loop performs $S/L-1$ rounds of the global gradient update. In the inner loop, $L$ rounds of minibatch stochastic gradient calculations are performed. When $b = 1$, the inner loop degrades into simple stochastic gradient descent. In the $l$-th iteration of the inner loop, the stochastic gradient is computed as follows:
\begin{equation}\label{18}
\setlength{\abovedisplayskip}{3pt}
\setlength{\belowdisplayskip}{3pt}
\boldsymbol{g}_l=\frac{1}{b}\sum_{i\in I}^{\,\,}{\left( \nabla \delta _i\left( \boldsymbol{\xi }_{sL+l} \right) -\nabla \delta _i\left( \boldsymbol{\omega } \right) \right)}+g,
\end{equation}
where $g$ is the gradient value obtained by traversing the entire dataset in the outer loop, and $\frac{1}{b}\sum_{i\in I}^{\,\,}{\left( \nabla \delta _i\left( \boldsymbol{\xi }_{sL+l} \right) -\nabla \delta _i\left( \boldsymbol{\omega } \right) \right)}$ is used to correct the gradient. Clearly, at the beginning of the inner loop, the variance of the estimate $\frac{1}{b}\sum_{i\in I}^{\,\,}{\left( \nabla \delta _i\left( \boldsymbol{\xi }_{sL+l} \right) -\nabla \delta _i\left( \boldsymbol{\omega } \right) \right)}$ is small, and thus the variance in estimating the entire gradient is also small. As $l$ increases, the iterations of the inner loop continue and the variance increases, and then the inner loop is jumped to reset $\boldsymbol{\omega }$ to estimate the new gradient value for training. The number of inner loops can be flexibly adjusted according to different convergence rate and computational complexity constraint.

Algorithm 1 can reduce computational complexity by using stochastic gradient rather than traversing the entire dataset in each iteration. In the HMC sampling method, it requires $S\cdot \left( NF \right) ^2$ gradient computations to get $S$ samples. However, our proposed SVRG-HMC sampling method only needs to traverse the entire dataset in the outer loop and use stochastic gradient in the inner loop, which just requires $S\cdot \left( NF + bL \right)$ gradient computations to get $S$ samples. Therefore, our proposed sampling method reduces $S\left( SN^2F^2-NF-bL \right) $ gradient computations, which achieves lower computational complexity.

\subsection{Global model integration phase}
In this subsection, we propose a communications-efficient FL based global model integration method. Firstly, we introduce a sampling method that simply combines Bayesian learning and FL. However, this native method will lead to much communications overhead in model integration. Secondly, we introduce the gradient quantization to achieve tradeoff between communications overhead and prediction accuracy. Thirdly, we adopt an encoding scheme for transmitting quantized gradient. Finally, we propose a communications-efficient global model integration for federated Bayesian learning sampling method.
\subsubsection{Native federated Bayesian learning sampling method}
The content popularity prediction method based on Bayesian learning can carry out model training and prediction on a single F-AP. However, the limitation of this approach is the small coverage of a single F-AP and the constraints of computing and caching resources. To solve this problem and take advantage of the huge amount of data in F-RANs, cooperative prediction participated by neighboring F-APs needs to be considered\cite{wuyuting}.

From Eq. (\ref{12}), we can infer that  $\boldsymbol{K}'$ and $\boldsymbol{\lambda }$ are related to the dimension of the local input dataset and their computational cost in Eq. (\ref{13}) is fixed. Let $\mathcal{R} _m$ denote the local dataset of the $m$-th F-AP. Then, the global posterior can be written as follows:
\begin{equation}\label{19}
\setlength{\abovedisplayskip}{3pt}
\setlength{\belowdisplayskip}{3pt}
p\left( \boldsymbol{\lambda },\boldsymbol{\beta }\left| \mathcal{R} \right. \right) \propto p\left( \boldsymbol{\lambda },\boldsymbol{\beta } \right) \prod_{m=1}^M{p\left( \mathcal{R} _m\left| \boldsymbol{\lambda },\boldsymbol{\beta } \right. \right)}.
\end{equation}
According to the previous definitions: $\rho _q=\log \left( \beta _q \right) $, $\boldsymbol{\tau }=\left[ \boldsymbol{\lambda }^T,\rho _0,...,\rho _{Q+1} \right] ^T\in R^{\left( F+Q+2 \right)}$, Eq. (\ref{19}) can be written as follows:
\begin{equation}\label{20}
\setlength{\abovedisplayskip}{1pt}
\setlength{\belowdisplayskip}{3pt}
p\left( \boldsymbol{\tau }\left| \mathcal{R} \right. \right) \propto p\left( \boldsymbol{\tau } \right) \prod_{m=1}^M{p\left( \mathcal{R} _m\left| \boldsymbol{\tau } \right. \right)}.
\end{equation}
Therefore, sampling from the global posterior can be naturally considered as a federated Bayesian learning sampling problem\cite{QLSD}. By utilizing our proposed SVRG-HMC sampling method in each F-AP, the sampling updates at the cloud server can be expressed as follows:
\begin{equation}\label{21}
\setlength{\abovedisplayskip}{1pt}
\setlength{\belowdisplayskip}{3pt}
\boldsymbol{\theta }_{k+1}=\left( 1-Dh \right) \boldsymbol{\theta }_{k+1}-hH_{k+1}\left( \boldsymbol{\tau }_k \right) +\sqrt{2Dh}\cdot \boldsymbol{\eta }_{k+1},
\quad\boldsymbol{\tau }_{k+1}=\boldsymbol{\tau }_k+h\boldsymbol{\theta }_{k+1},
\end{equation}
where $H_{k+1}\left( \boldsymbol{\tau }_k \right) $ represents the unbiased estimate of the gradient $\nabla \log p\left( \boldsymbol{\tau }|\mathcal{R} \right) $.

In the FL framework, the $m$-th F-AP obtains the gradient estimate $H_{k+1}^{\left( i \right)}$ at the $k$-th iteration based on the local dataset $\mathcal{R} _i$. Then, $H_{k+1}=\sum_{i=1}^b{H_{k+1}^{\left( i \right)}}$ is the gradient estimate of the whole dataset. Also, we assume that $H_{k+1}$ is the unbiased estimate of the global gradient while $H_{k+1}^{\left( i \right)}$ is not an unbiased estimator, which helps to use a biased local stochastic gradient with better convergence guarantee\cite{QLSD}. Then, Eq. (\ref{21}) can be written as follows:
\begin{equation}\label{22}
\setlength{\abovedisplayskip}{1pt}
\setlength{\belowdisplayskip}{3pt}
\boldsymbol{\theta }_{k+1}=\left( 1-Dh \right) \boldsymbol{\theta }_{k+1}-\frac{h}{M}\sum_{m=1}^M{H_{k+1}^{\left( i \right)}}\left( \boldsymbol{\tau }_k \right)+\sqrt{2Dh}\cdot\boldsymbol{\eta }_{k+1},
\quad\boldsymbol{\tau }_{k+1}=\boldsymbol{\tau }_k+h\boldsymbol{\theta }_{k+1}.
\end{equation}

By applying the model training method of a single F-AP, the federated Bayesian learning sampling algorithm is presented in Algorithm 2. In Algorithm 2, each F-AP computes gradients based on its local dataset and sends them to the cloud server, and the cloud server then performs federated averaging and updates the samples. It integrates FL and Bayesian learning effectively. The reason is that the flexibility of FL allows each F-AP to generate local model through Bayesian learning, and FL further exploits the computing resources of each F-AP and the data in F-RANs together.
\begin{algorithm}[!t]
\algsetup{linenosize=\scriptsize} \footnotesize
\caption{ Native federated Bayesian learning sampling method. }
\begin{algorithmic}[1]
\REQUIRE $S, L, b, h>0, Dh<1, D\geqslant 1$;
\ENSURE $\left\{ \boldsymbol{\tau }_s \right\} _{1\leqslant s\leqslant S/L-1}$;
\STATE Initialize $\left\{ \boldsymbol{\theta }_0,\boldsymbol{\theta }_{0}^{\left( 1 \right)},...,\boldsymbol{\theta }_{0}^{\left( b \right)} \right\} , \left\{ \boldsymbol{\tau }_0,\boldsymbol{\tau }_{0}^{\left( 1 \right)},...,\boldsymbol{\tau }_{0}^{\left( b \right)} \right\} $;

\FOR {$s=0,1,...,S$}
\STATE Choose participants $\mathcal{A} _{s+1}$;

\FOR {$i\in \mathcal{A} _{s+1}$}
\STATE (in parallel over nodes)
\STATE $g=-\frac{1}{NF}\sum\nolimits_{i=1}^{NF}{\nabla \log p\left( r_{c_f}\left[ n \right] |\lambda _f \right)}=\frac{1}{NF}\sum\nolimits_{i=1}^{NF}{\nabla \delta _i\left( \boldsymbol{\xi } \right)}$;
\STATE $\boldsymbol{\omega }=\boldsymbol{\tau }_{s}^{\left( i \right)}$;
\FOR {$l=0,1,...,L-1$}
\STATE Randomly draw a subset $I\subseteq \mathcal{R} , \left| I \right|=b$ with equal probability;
\STATE $\tilde{\nabla}_{s}^{\left( i \right)}=-\log p\left( \boldsymbol{\tau }_{s}^{\left( i \right)} \right) -\sum_{q=0}^{Q+1}{\log p\left( \boldsymbol{\tau }_{s}^{\left( i \right)} \right)}+\frac{NF}{b}\sum_{i\in I}^{\,\,}{\left( \nabla \delta _i\left( \boldsymbol{\tau }_{s}^{\left( i \right)} \right) -\nabla \delta _i\left( \boldsymbol{w} \right) \right)}+g$;
\STATE Send $\tilde{\nabla}_{s,i}$ to the cloud server;
\ENDFOR
\ENDFOR
\STATE Federated averaging $\tilde{\nabla}_s=\frac{b}{\left| \mathcal{A} _{s+1} \right|}\sum\nolimits_{i\in \mathcal{A} _{k+1}}^{\,\,}{\tilde{\nabla}_{s}^{\left( i \right)}}$;
\STATE Draw $\boldsymbol{\eta }_{s+1}\sim \mathcal{N} \left( 0,\boldsymbol{I}_d \right) $;
\STATE $\boldsymbol{\theta }_{s+1}=\left( 1-Dh \right) \boldsymbol{\theta }_s-h\tilde{\nabla}_s+\sqrt{2Dh}\cdot \boldsymbol{\eta }_{s+1}$;
\STATE $\boldsymbol{\tau }_{s+1}=\boldsymbol{\tau }_s+h\boldsymbol{\theta }_{s+1}$;
\STATE Send $\boldsymbol{\tau }_{s+1}$ to $M$ F-APs;
\ENDFOR
\end{algorithmic}
\end{algorithm}

\subsubsection{Gradient quantization}
If the gradient vectors are dense, each participant needs to send and receive multiple floating point numbers to each other in each iteration to update the parameter vectors. In practical applications, transmitting gradients in each iteration is a significant performance bottleneck.

In this phase, we adopt a gradient quantization scheme. Let $\boldsymbol{v}\in \mathbb{R} ^d$ denote the gradient vector passed in FL,  $\left\| \boldsymbol{v} \right\| _0$ the number of nonzero elements in $\boldsymbol{v}$ and $\left\| \boldsymbol{v} \right\| _2=\sqrt{v_{1}^{2}+v_{2}^{2}...+v_{d}^{2}}$. The number of bits required to encode a floating number is denoted by $F$ (typically 32 or 64 bits, considered to be a constant in the following). For any binary code string $\omega$, let $\left| \omega \right|$ denote its length. For any scalar, let $\mathrm{sgn} \left( x \right) \in \left\{ -1,+1 \right\} $ denote its sign and $\mathrm{sgn} \left( 0 \right)=1$.

We define a random quantization function such that for any $d$-dimensional vector $\boldsymbol{v}\in \mathbb{R} ^d$ with $\boldsymbol{v}\ne \mathbf{0}$, the $i$-th element of its quantized vector $Q\left( \boldsymbol{v} \right) $ can be expressed as $Q_i\left( \boldsymbol{v},s \right) =\left\| \boldsymbol{v} \right\| _2\cdot \sgn \left( v_i \right) \xi _i\left( \boldsymbol{v},s \right)$\cite{QSGD},
where $s$ is a quantization level, and $\xi _i\left( \boldsymbol{v},s \right) $ is an independent random variable. Let $0\leqslant l_i<s$ be a positive integer such that $\frac{\left| v_i \right|}{\,\,\left\| \boldsymbol{v} \right\| _2}\in \left[ \frac{l_i}{s},\frac{l_i+1}{s} \right] $. Then, $\xi _i\left( \boldsymbol{v},s \right) $ can be defined as follows:
\begin{equation}\label{24}
\setlength{\abovedisplayskip}{1pt}
\setlength{\belowdisplayskip}{3pt}
\xi _i\left( \boldsymbol{v} \right) =\begin{cases}	\,\, \frac{l_i}{s}\,\,   \qquad\quad           p=1-q\left( \frac{\left| v_i \right|}{\,\,\left\| \boldsymbol{v} \right\| _2},s \right)\\	\frac{l_i+1}{s}\,\,          \,\,\quad\quad       p=q\left( \frac{\left| v_i \right|}{\,\,\left\| \boldsymbol{v} \right\| _2},s \right)\\\end{cases},
\end{equation}
where $q\left( \frac{\left| v_i \right|}{\,\,\left\| \boldsymbol{v} \right\| _2},s \right) =\frac{s\cdot \left| v_i \right|}{\,\,\left\| \boldsymbol{v} \right\| _2}-l_i$. According to the relationship of $\frac{\left| v_i \right|}{\,\,\left\| \boldsymbol{v} \right\| _2}\in \left[ \frac{l_i}{s},\frac{l_i+1}{s} \right] $, we can establish $q\left( \frac{\left| v_i \right|}{\,\,\left\| \boldsymbol{v} \right\| _2},s \right) \in \left[ 0,1 \right]$, which satisfies the definition of probability distribution. Then, we have the following lemma.

\newtheorem{lemma}{Lemma}[section]
\begin{lemma}
For any $\boldsymbol{v}\in \mathbb{R} ^d$, we have:
\vspace{-1em}
\begin{align}
\setlength{\abovedisplayskip}{1pt}
\setlength{\belowdisplayskip}{3pt}
\mathbb{E} \left[ \left\| Q\left( \boldsymbol{v},s \right) \right\| _0 \right] &\leqslant \sqrt{d}+s^2 ,\\
\mathbb{E} \left[ Q\left( \boldsymbol{v},s \right) \right] &=\boldsymbol{v},\\
\mathbb{E} \left[ \left\| Q\left( \boldsymbol{v},s \right) \right\| _{2}^{2} \right] &\leqslant \left( 1+\min \left( \frac{d}{\,\,s^2},\frac{\sqrt{d}}{s} \right) \right) \left\| \boldsymbol{v} \right\| _{2}^{2}.
\end{align}
\label{spar}
\end{lemma}
\vspace{-2.5em}
\begin{proof}
Please see Appendix A.
\end{proof}
Lemma \ref{spar} demonstrates the sparsity, unbiasedness and the upper bound on the estimated variance of $Q\left( \boldsymbol{v},s \right)$.

\begin{figure}[b!]
\vspace{-2em}
\centering
\includegraphics[width=0.6\linewidth,height=0.08\linewidth]{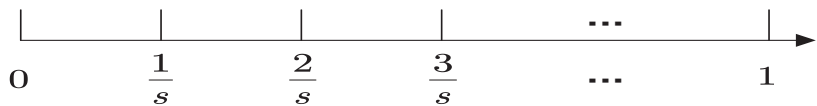}
\caption{Illustration of the quantization level $s$.}
\label{MeanOfS}
\end{figure}

It can be observed that the properties of $Q\left( \boldsymbol{v},s \right)$ is determined by $s$. As shown in Fig. \ref{MeanOfS}, the purpose of $s$ is to provide a higher quantization accuracy, which is equivalent to dividing the quantization result into discrete $s$ points from the coordinate. For each $v_i$, the corresponding $l_i$ is selected so that $\frac{\left| v_i \right|}{\,\,\left\| \boldsymbol{v} \right\| _2}$ falls within the interval $\left[ \frac{l_i}{s},\frac{l_i+1}{s} \right] $. The left and right endpoints of the interval are two quantization levels, and the larger $s$, the smaller the corresponding quantization accuracy $\frac{1}{s}$ and the more accurate the result. In the calculation, a larger $\frac{\left| v_i \right|}{\,\,\left\| \boldsymbol{v} \right\| _2}$ leads to a greater probability of being quantized to the right endpoint of the interval $\frac{l_i+1}{s}$; similarly, a smaller $\frac{\left| v_i \right|}{\,\,\left\| \boldsymbol{v} \right\| _2}$ leads to a greater probability of being quantized to the left endpoint of the interval $\frac{l_i}{s}$, which is consistent with the intuitive perception of quantization.

From Lemma \ref{spar}, we can infer that the upper bound of $\mathbb{E} \left[ \left\| Q\left( \boldsymbol{v},s \right) \right\| _{2}^{2} \right] $ is affected by the vector dimension $d$ and the quantization level $s$. When $s=1$, the upper bound of $\mathbb{E} \left[ \left\| Q\left( \boldsymbol{v},s \right) \right\| _{2}^{2} \right] $ is $\left( 1+\sqrt{d} \right) \left\| \boldsymbol{v} \right\| _{2}^{2}$. When $s$ is adjusted from 1 to $\sqrt{d}$, the upper bound of $\mathbb{E} \left[ \left\| Q\left( \boldsymbol{v},s \right) \right\| _{2}^{2} \right] $ varies between $O\left( \sqrt{d} \right) $ and $O\left( 1 \right) $. It can be observed that the sparsity of the quantized gradient decreases as $s$ increases. Therefore, for a given gradient vector dimension, a suitable quantization level $s$ needs to be specified for quantization to obtain a tradeoff between prediction accuracy and transmission efficiency. Then, we have the following theorem.

\newtheorem{lemma2}{Lemma}[section]
\newtheorem{theorem}{Theorem}[section]
\begin{theorem}
For any $\boldsymbol{v}\in \mathbb{R} ^d$ and $s^2+\sqrt{d}\leqslant \frac{d}{2}$, there exists a coding scheme which allows the upper bound on the number of bits required to transmit $Q\left( \boldsymbol{v} ,s \right) $ to be expressed as:
\begin{equation}
F+\left( 3+\frac{3}{2}\cdot \left( 1+O\left( 1 \right) \right) \log _2\left( \frac{2\left( s^2+d \right)}{s^2+\sqrt{d}} \right) \right) \left( s^2+\sqrt{d} \right).
\end{equation}
\label{UpperBound}
\end{theorem}
\vspace{-2.5em}
\begin{proof}
Please see Appendix B.
\end{proof}

Theorem \ref{UpperBound} shows the upper bound of communications overhead for transmitting $Q\left( \boldsymbol{v} ,s \right) $. In summary, for any gradient $\boldsymbol{v}$, its quantizer output $Q\left( \boldsymbol{v},s \right) $ can be represented by a tuple $\left( \left\| \boldsymbol{v} \right\| _2,\boldsymbol{\gamma },\boldsymbol{\varepsilon } \right) $, where $\boldsymbol{\gamma }$ is the vector representing the sign of each element $v_i$ of $\boldsymbol{v}$, and $\boldsymbol{\varepsilon }$ is the vector representing the integer $s\cdot \xi _i\left( \boldsymbol{v},s \right) $. By encoding these messages, the communications overhead can be greatly reduced.

\subsubsection{Coding scheme}
In practical applications, the value of $s$ is small to maintain sparsity in Lemma \ref{spar}, so the element $s\cdot \xi _i\left( \boldsymbol{v},s \right) $ in $\boldsymbol{\varepsilon }$ is usually not a big integer\cite{QLSD}. If a fixed-length encoding scheme is used to represent these values, each of them requires a full 32-bit integer, but the number of their effective bits is usually only 10 or even less, resulting in a great waste of communications resources. Therefore, when the quantization results are transmitted, a suitable variable-length encoding scheme is required.

For any gradient $\boldsymbol{v}$, the quantizer output $Q\left( \boldsymbol{v},s \right) $ can be represented by a tuple $\left( \left\| \boldsymbol{v} \right\| _2,\boldsymbol{\gamma },\boldsymbol{\varepsilon } \right) $. Let $Q\left( \boldsymbol{v},s \right) $ denote a function that maps from $\mathbb{R} ^d$ to $\mathcal{B} _s$, where $\mathcal{B} _s$ is defined as follows:
\begin{equation}\label{26}
\setlength{\abovedisplayskip}{1pt}
\setlength{\belowdisplayskip}{3pt}
\mathcal{B} _s=\left\{ \left( A,\boldsymbol{\gamma },\boldsymbol{y} \right): A\in \mathbb{R},\gamma _i\in \left\{ -1,+1 \right\},y_i\in \left\{ 0,\frac{1}{s},...,1 \right\} \right\} .
\end{equation}
Note that the tuple $\mathcal{B} _s$ contains all the information of the gradient $\boldsymbol{v}$. Therefore, we can encode $\mathcal{B} _s$ instead of $\boldsymbol{v}$.

There are some classic variable-length encoding schemes such as Shannon coding\cite{shannon} and Huffman coding\cite{huffman}. However, both of them need to know the appearing probability of each integer in advance, which is not applicable to our proposed quantization scheme. Instead, Elias coding is an efficient method to encode integer\cite{Elias}. The procedure of Elias coding is presented in Algorithm 3.
\begin{algorithm}[t]
\algsetup{linenosize=\scriptsize} \footnotesize
\caption{ Elias coding method. }
\begin{algorithmic}[1]
\REQUIRE Integer $k$;
\ENSURE Coding string $\mathrm{Elias}\left( k \right)$;
\STATE Add a zero to the right of the coding string;
\WHILE {$\lfloor \log _2k \rfloor >0$}
\STATE Write the binary encoding result of $k$ on the left of the coding string;
\STATE $k\gets \lfloor \log _2k \rfloor $;
\ENDWHILE
\end{algorithmic}
\end{algorithm}
For example, $\mathrm{Elias}\left( 1 \right) =0$, $\mathrm{Elias}\left( 8 \right) =1110000$. The decoding step simply involves an inverse transformation from left to right of $\mathrm{Elias}\left( k \right)$ to determine the original value. Then, we have the following lemma which describes a well-known property of Elias coding\cite{EliasProperty}.
\newtheorem{lemma3}{Lemma}[section]
\begin{lemma}
For any positive integer $k$:
\vspace{-1em}
\begin{align}\label{27}
\left| \mathrm{Elias}\left( k \right) \right|\leqslant \log _2k+\log _2\left( \log _2k \right)+\cdots+1=\left( 1+O\left( 1 \right) \right) \log _2k+1.
\end{align}
\label{LengthOfElias}
\end{lemma}
\vspace{-2.5em}
Lemma \ref{LengthOfElias} shows the number of bits required to encode a positive integer $k$ by Elias coding.

Given a tuple $\left( A,\boldsymbol{\gamma },\boldsymbol{y} \right) \in \mathcal{B} _s$, let $\mathrm{Code}\left( A,\boldsymbol{\gamma },\boldsymbol{y} \right) $ denote the number of bits required for transmitting the tuple, where $A$ is a floating number which requires $F$ bits to encode. Then, the Elias coding scheme can be described as follows. First, the location of the first non-zero coordinate in $\boldsymbol{y}$ is encoded using Elias coding. After that, a bit representing $\gamma _i$ is appended and followed by the $\mathrm{Elias}\left( sy_i \right) $. Let $c\left( y_i \right) $ denote the distance from current coordinate of $\boldsymbol{y}$ to the next non-zero coordinate $y_{i-\mathrm{next}}$. Iteratively, encode $c\left( y_i \right) $ by Elias coding. $\gamma _{i-\mathrm{next}}$ and $y_{i-\mathrm{next}}$ of that coordinate are encoded by Elias coding in the same way. The decoding steps are similar: first read the $F$ bits to reconstruct $A$, and then iteratively use the decoding scheme for Elias coding to read off the positions and values of the non-zero elements of $\boldsymbol{\gamma }$ and $\boldsymbol{y}$.

By using the Elias encoding scheme, bit waste can be effectively reduced. Therefore, introducing Elias coding into our federated Bayesian learning sampling framework is another important measure and innovation to reduce the communications overhead.

\subsubsection{Communications-efficient federated Bayesian learning sampling method}
In this subsection, we propose a communications-efficient federated Bayesian learning sampling method. The proposed sampling method is presented in Algorithm 4. By comparing Algorithm 4 with Algorithm 2, it can be found that our proposed communications-efficient federated Bayesian learning sampling method adds random quantization and encoding steps. Each F-AP still calculates the gradient based on the current local dataset, then quantizes and encodes the gradient vector and sends it to the cloud server. In Algorithm 4, the random quantization will introduce estimation error. The main idea is to use $\left[ 0,\left\| \boldsymbol{v} \right\| _2 \right] $ as the quantization interval of $v_i$ in the gradient vector, then divide the entire interval into $s$ parts, and quantize each element to the corresponding interval endpoints. The intuition is that a gradient vector $\boldsymbol{v}$ can be expressed by $\left( \left\| \boldsymbol{v} \right\| _2,\boldsymbol{\gamma },\boldsymbol{ \varepsilon } \right) $ and its elements represent the magnitude, sign and quantization level of the vector, respectively. Since the quantization level is an integer, it can be transmitted by Elias encoding, which can improve communications efficiency. In this way, although estimation error is introduced to the gradient vector, the communications overhead can be effectively reduced. The bound of estimation error has been shown in Lemma \ref{spar}, which decreases gradually with the increase of $s$.
\begin{algorithm}[t!]
\algsetup{linenosize=\scriptsize} \footnotesize
\caption{Communications-efficient federated Bayesian learning sampling method.}
\begin{algorithmic}[1]
\REQUIRE $S, L, b, h>0, Dh<1, D\geqslant 1$;
\ENSURE $\left\{ \boldsymbol{\tau }_s \right\} _{1\leqslant s\leqslant S/L-1}$;
\STATE Initialize $\left\{ \boldsymbol{\theta }_0,\boldsymbol{\theta }_{0}^{\left( 1 \right)},...,\boldsymbol{\theta }_{0}^{\left( b \right)} \right\} , \left\{ \boldsymbol{\tau }_0,\boldsymbol{\tau }_{0}^{\left( 1 \right)},...,\boldsymbol{\tau }_{0}^{\left( b \right)} \right\} $;

\FOR {$s=0,1,...,S$}
\STATE Choose participants $\mathcal{A} _{s+1}$;

\FOR {$i\in \mathcal{A} _{s+1}$}
\STATE (in parallel over nodes)
\STATE $g=-\frac{1}{NF}\sum\nolimits_{i=1}^{NF}{\nabla \log p\left( r_{c_f}\left[ n \right] |\lambda _f \right)}=\frac{1}{NF}\sum\nolimits_{i=1}^{NF}{\nabla \delta _i\left( \boldsymbol{\xi } \right)}$;
\STATE $\boldsymbol{\omega }=\boldsymbol{\tau }_{s}^{\left( i \right)}$;
\FOR {$l=0,1,...,L-1$}
\STATE Randomly draw a subset $I\subseteq \mathcal{R} , \left| I \right|=b$ with equal probability;
\STATE $\tilde{\nabla}_{s}^{\left( i \right)}=-\log p\left( \boldsymbol{\tau }_{s}^{\left( i \right)} \right) -\sum_{q=0}^{Q+1}{\log p\left( \boldsymbol{\tau }_{s}^{\left( i \right)} \right)}+\frac{NF}{b}\sum_{i\in I}^{\,\,}{\left( \nabla \delta _i\left( \boldsymbol{\tau }_{s}^{\left( i \right)} \right) -\nabla \delta _i\left( \boldsymbol{w} \right) \right)}+g$;
\STATE Quantize and encode to obtain $\mathscr{L} \left( \tilde{\nabla}_{s}^{\left( i \right)} \right) $;
\STATE Send $\mathscr{L} \left( \tilde{\nabla}_{s}^{\left( i \right)} \right) $ to the cloud server;
\ENDFOR
\ENDFOR
\STATE The cloud server decodes $\mathscr{L} \left( \tilde{\nabla}_{s}^{\left( i \right)} \right) $ to obtain $\tilde{\nabla}_{s}^{\left( i \right) *}$;
\STATE Federated averaging $\tilde{\nabla}_s=\frac{b}{\left| \mathcal{A} _{s+1} \right|}\sum\nolimits_{i\in \mathcal{A} _{k+1}}^{\,\,}{\tilde{\nabla}_{s}^{\left( i \right) *}}$;
\STATE Draw $\boldsymbol{\eta }_{s+1}\sim \mathcal{N} \left( 0,\boldsymbol{I}_d \right) $;
\STATE $\boldsymbol{\theta }_{s+1}=\left( 1-Dh \right) \boldsymbol{\theta }_s-h\tilde{\nabla}_s+\sqrt{2Dh}\cdot \boldsymbol{\eta }_{s+1}$;
\STATE $\boldsymbol{\tau }_{s+1}=\boldsymbol{\tau }_s+h\boldsymbol{\theta }_{s+1}$;
\STATE Send $\boldsymbol{\tau }_{s+1}$ to $M$ F-APs;
\ENDFOR
\end{algorithmic}
\end{algorithm}

According to Algorithm 4, we can collect enough samples to approximate the posterior distribution. As the number of sampling rounds increases, the collected samples will be closer to the true distribution. The initial samples are usually discarded, which are called burn-in samples. After collecting enough samples, they can be used for the subsequent predicting phase.

\subsection{Predicting phase}
Assuming that $N$ time periods have been observed, we demonstrate prediction for two cases. The first case is to predict the number of requests for existing contents in the content library and the second case is for newly-added contents.

For the first case, we can derive the distribution of new requests as follows:
\begin{equation}\label{28}
\setlength{\abovedisplayskip}{1pt}
\setlength{\belowdisplayskip}{3pt}
p\left( \boldsymbol{r}_c\left[ N+1 \right]\left| \mathcal{R} \right. \right) =\int{p\left(\boldsymbol{r}_c\left[ N+1 \right]\left| \boldsymbol{\lambda } \right. \right) p\left( \boldsymbol{\lambda }\left| \mathcal{R} \right. \right)}{\rm{d}}\boldsymbol{\lambda }.
\end{equation}
According to our proposed probabilistic model, $p\left( \boldsymbol{r}_c\left[ N+1 \right]\left| \boldsymbol{\lambda } \right. \right) $ is a Poisson distribution and $p\left( \boldsymbol{\lambda }\left| \mathcal{R} \right. \right) $ is a marginal likelihood function of $\boldsymbol{\lambda }$ based on the request observations $\mathcal{R}$. Nevertheless, it is often intractable to derive the integral operation. Instead, we make a point prediction rather than estimate the entire posterior distribution\cite{bayesianchoice}. We define a quadratic loss function to measure the loss between the predicted and actual values. Minimizing the loss function, we can obtain the mean of the predictive distribution and approximate it as follows:
\begin{equation}\label{29}
\setlength{\abovedisplayskip}{1pt}
\setlength{\belowdisplayskip}{3pt}
\mathbb{E}\left( \boldsymbol{r}_c\left[ N+1 \right]\left| \mathcal{R} \right. \right) \approx \frac{1}{S}\sum_{s=1}^S{e^{\boldsymbol{\lambda }^{\left( s \right)}}}.
\end{equation}

For the second case, the posterior distribution is defined as follows:
\begin{equation}\label{30}
\setlength{\abovedisplayskip}{1pt}
\setlength{\belowdisplayskip}{3pt}
p\left( \lambda _{F+1}\left| \boldsymbol{x}_{F+1} \right. \right) =\int{p\left( \lambda _{F+1}\left| \boldsymbol{\lambda },\boldsymbol{\beta },\boldsymbol{x}_{F+1} \right. \right)}\cdot p\left( \boldsymbol{\lambda },\boldsymbol{\beta }\left| \mathcal{R} \right. \right) {\rm{d}}\boldsymbol{\beta }{\rm{d}}\boldsymbol{\lambda },
\end{equation}
where $\boldsymbol{x}_{F+1}$ represents the feature vector of a newly-added content. Note that the joint distribution of $p\left( \lambda _1,\lambda _2,\dots ,\lambda _{F+1} \right) $ is a normal distribution with zero mean and covariance matrix:
\begin{equation}\label{31}
\setlength{\abovedisplayskip}{1pt}
\setlength{\belowdisplayskip}{3pt}
\left[ \begin{matrix}	\,\,\boldsymbol{K}'&		\boldsymbol{k}'\\	\boldsymbol{k}'^T&		K\left( \boldsymbol{x}_{F+1},\boldsymbol{x}_{F+1} \right)\\\end{matrix} \right],
\end{equation}
where $\boldsymbol{k}'=\left[ K\left( \boldsymbol{x}_1,\boldsymbol{x}_{F+1} \right) ,\dots ,K\left( \boldsymbol{x}_F,\boldsymbol{x}_{F+1} \right) \right]^T $. According to the property of the Gaussian distribution, the conditional distribution $p\left( \lambda _{F+1}\left| \boldsymbol{\lambda },\boldsymbol{x}_{F+1},\boldsymbol{\beta } \right. \right) $ is a normal distribution with mean and variance:
\begin{equation}
\setlength{\abovedisplayskip}{1pt}
\setlength{\belowdisplayskip}{3pt}
\lambda _{F+1}=\boldsymbol{k}'^T\boldsymbol{K}'^{-1}\boldsymbol{\lambda },\quad\sigma _{F+1}=K\left( \boldsymbol{x}_{F+1},\boldsymbol{x}_{F+1} \right) -\boldsymbol{k}'^T\boldsymbol{K}'^{-1}\boldsymbol{k}'.
\end{equation}
Similarly, the point estimation of the request rate for a newly-added content can be approximated as follows:
\begin{equation}\label{32}
\setlength{\abovedisplayskip}{1pt}
\setlength{\belowdisplayskip}{3pt}
\mathbb{E}\left( r_{F+1}\left[ N+1 \right]\left| \boldsymbol{x}_{F+1} \right. \right) \approx \frac{1}{S}\sum_{s=1}^S{e^{\lambda _{F+1}^{\left( S \right)}+\frac{1}{2}\sigma _{F+1}^{\left( S \right)}}}.
\end{equation}

\begin{figure}[b!]
\vspace{-2em}
\centering
\includegraphics[width=0.9\linewidth,height=0.17\linewidth]{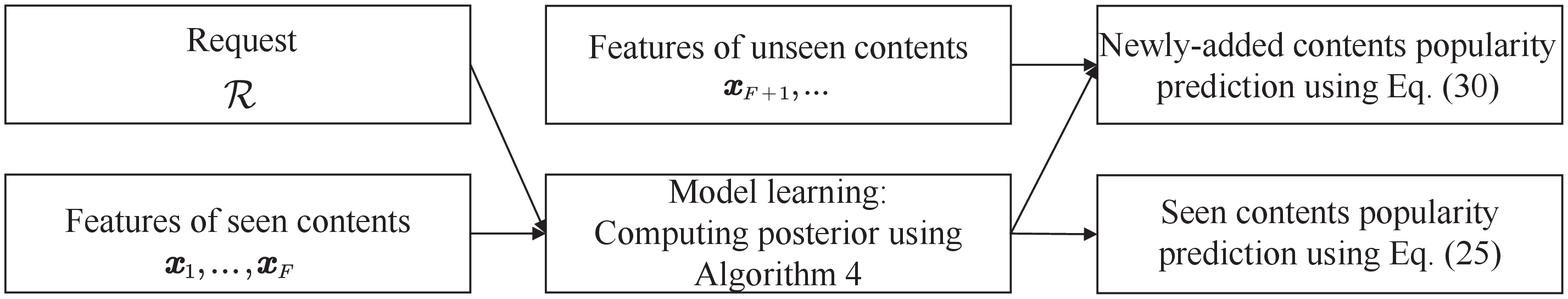}
\caption{Procedure of our proposed content popularity prediction policy.}
\label{procedure}
\vspace{-2em}
\end{figure}

According to the above descriptions, the procedure of our proposed popularity prediction policy is shown in Fig. \ref{procedure}. In the model learning phase, the model parameters are learned based on the number of requests and the features of contents stored in the cloud server (referred to as seen contents). In the predicting phase, the proposed policy enables to predict the popularities of both seen and newly-added contents (referred to as unseen contents). Besides, considering that content popularity can be a temporal variant, we can shorten the time period to track content popularity in peak hours.

\section{Simulation Results}
To evaluate the performance of our proposed popularity prediction policy, we take movie contents as an example and perform simulations using the data extracted from the MovieLens 100K Dataset\cite{movielens}. The dataset consists of 100,000 ratings (1-5) from 943 users on 1,682 movies. Each item contains the information about the movies including the features of rating, category and timestamp. We take the rating as the number of requests for the movies. Besides, we set the number of the considered F-APs $M$ to 5, the preset time period to twelve hours and the dimension of content feature vector to 10, respectively.

\begin{figure}[b!]
    \vspace{-2em}
	\centering
	\begin{minipage}{0.49\linewidth}
		\centering
		\includegraphics[width=0.9\linewidth,height=0.4\linewidth]{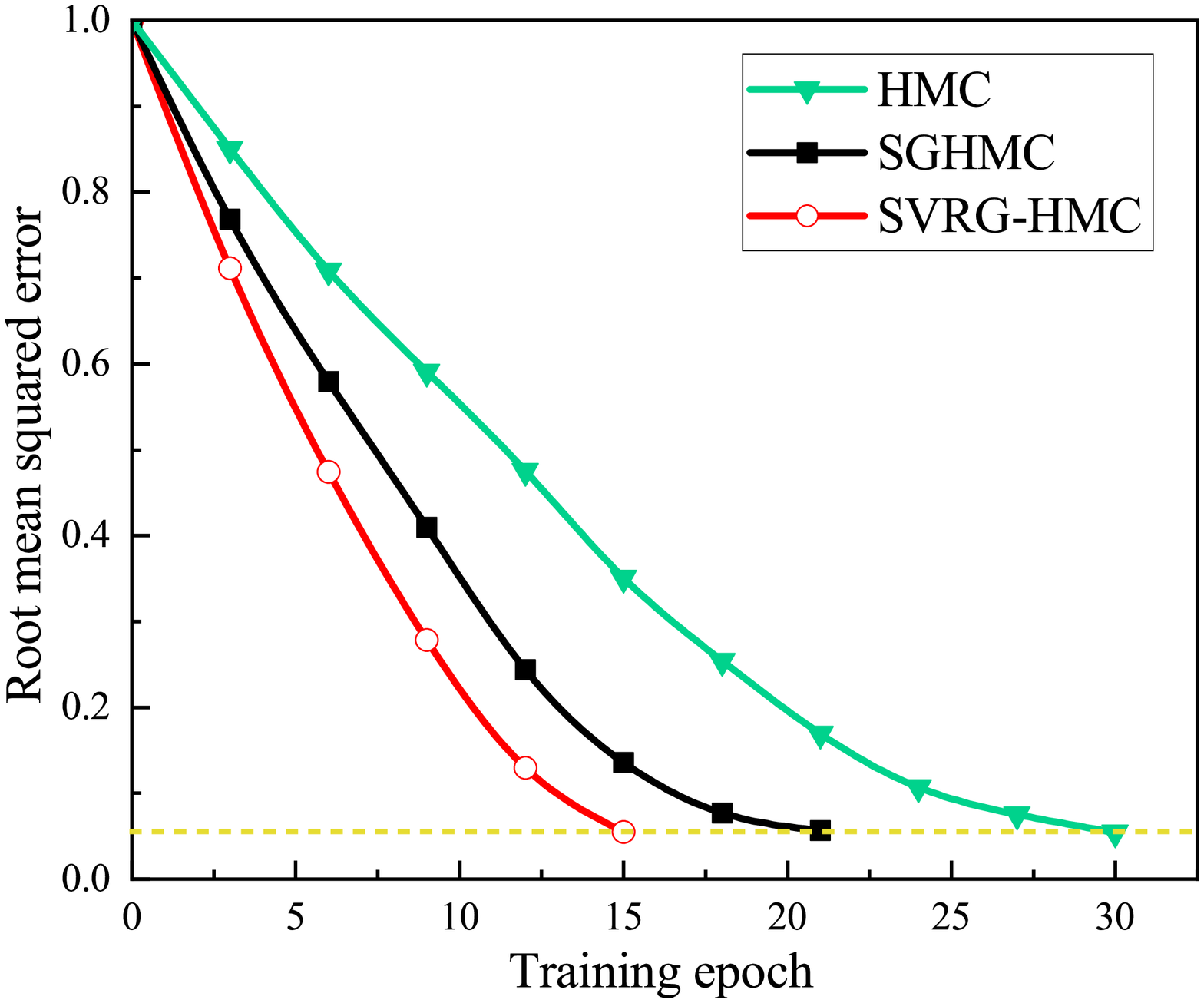}
		\caption{RMSE versus training epoch.}
		\label{svrgconv}
	\end{minipage}
	\begin{minipage}{0.49\linewidth}
		\centering
		\includegraphics[width=0.9\linewidth,height=0.4\linewidth]{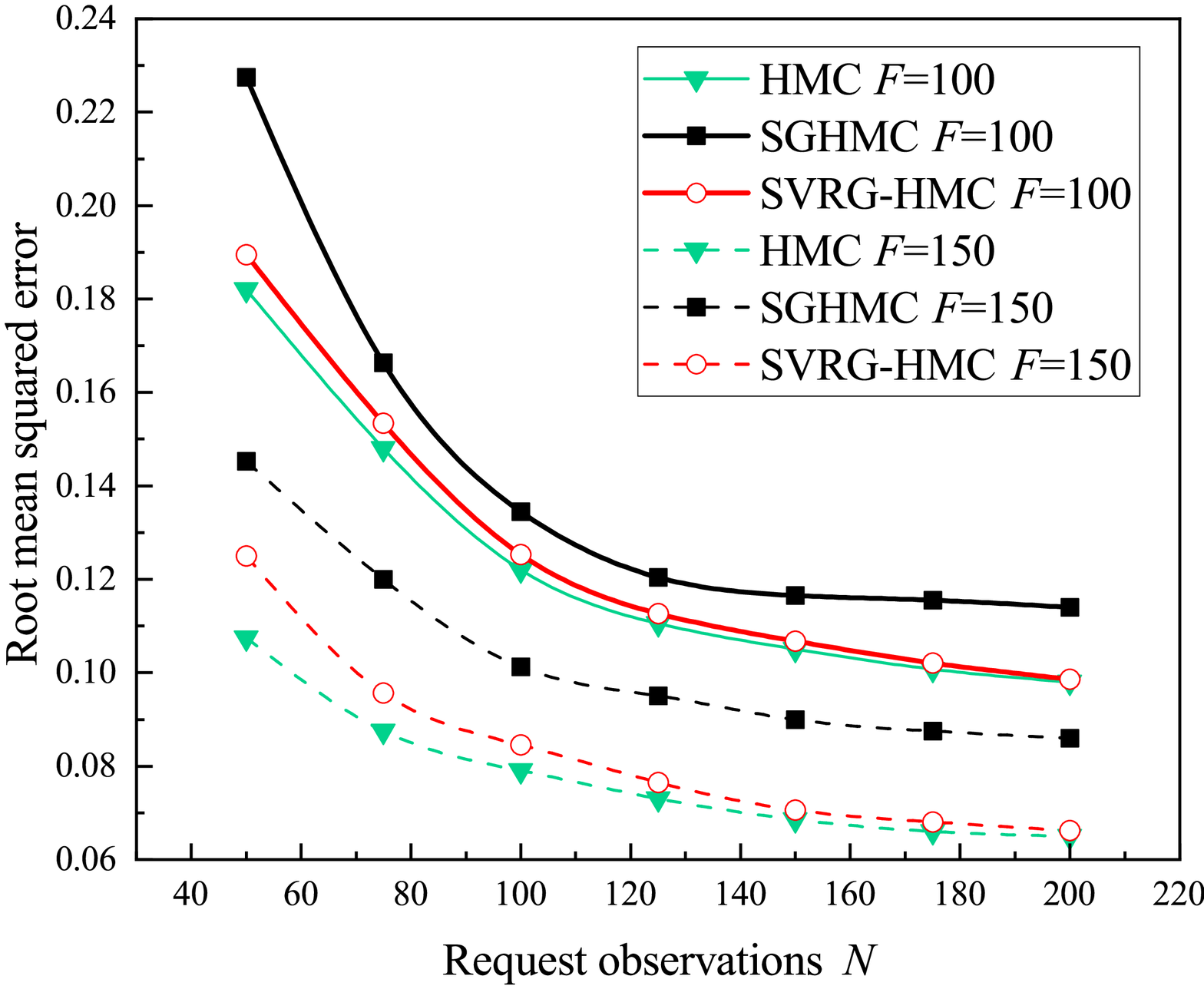}
		\caption{RMSE versus request observations with different $F$.}
		\label{svrggmc}
	\end{minipage}
\end{figure}

In order to validate our theoretical results, our simulation results include two parts. First, we compare our proposed Bayesian learning based prediction policy in local training phase with traditional Bayesian learning based prediction policy in Fig. \ref{svrgconv} and Fig. \ref{svrggmc}. These policies are performed on a single F-AP. The aim of this part is to validate the advantages of introducing SVRG to traditional sampling method in Bayesian learning. The reminder of simulation results is to show the performance of our proposed quantized federated Bayesian learning based content popularity prediction policy. 

In Fig. \ref{svrgconv}, we show the RMSE of HMC, stochastic gradient HMC (SGHMC) and SVRG-HMC based local content popularity prediction policy versus training epoch. These policies are performed at a single F-AP without aggregating. It can be observed that SVRG-HMC and SGHMC take less training epoches to converge to the same accuracy, and reach higher accuracies in shorter epoches, which can meet the need of real-time prediction of content popularity. The reason is that SVRG-HMC and SGHMC utilize stochastic gradients to iterate only a small portion of the whole dataset. Besides, The introduction of variance reduction also accelerates the convergence due to the simple use of stochastic gradient in SGHMC.

In Fig. \ref{svrggmc}, we show the RMSE of HMC, SGHMC and SVRG-HMC based local content popularity prediction policy versus request observations $N$. It can be observed that the RMSE of our proposed SVRG-HMC based local content popularity prediction policy and the other HMC based policies decreases as the number of request observations increases. The reason is that the Gaussian process in the probabilistic model allows for a better insight into the relationship between the number of requests and the content features by obtaining more observation samples. It can also be observed that the RMSE of SVRG-HMC is significantly smaller than that of SGHMC and close to that of HMC. The reason is that stochastic gradient introduces prediction error, and SVRG-HMC utilizes variance reduction to alleviate this impact. Specifically, it can be observed that the RMSE decreases when the number of contents in the library $F$ increases.

\begin{figure}[t!]
	\centering
	\begin{minipage}[b!]{0.5\linewidth}
        \vspace{0pt}
		\centering
		\includegraphics[width=0.9\linewidth,height=0.4\linewidth]{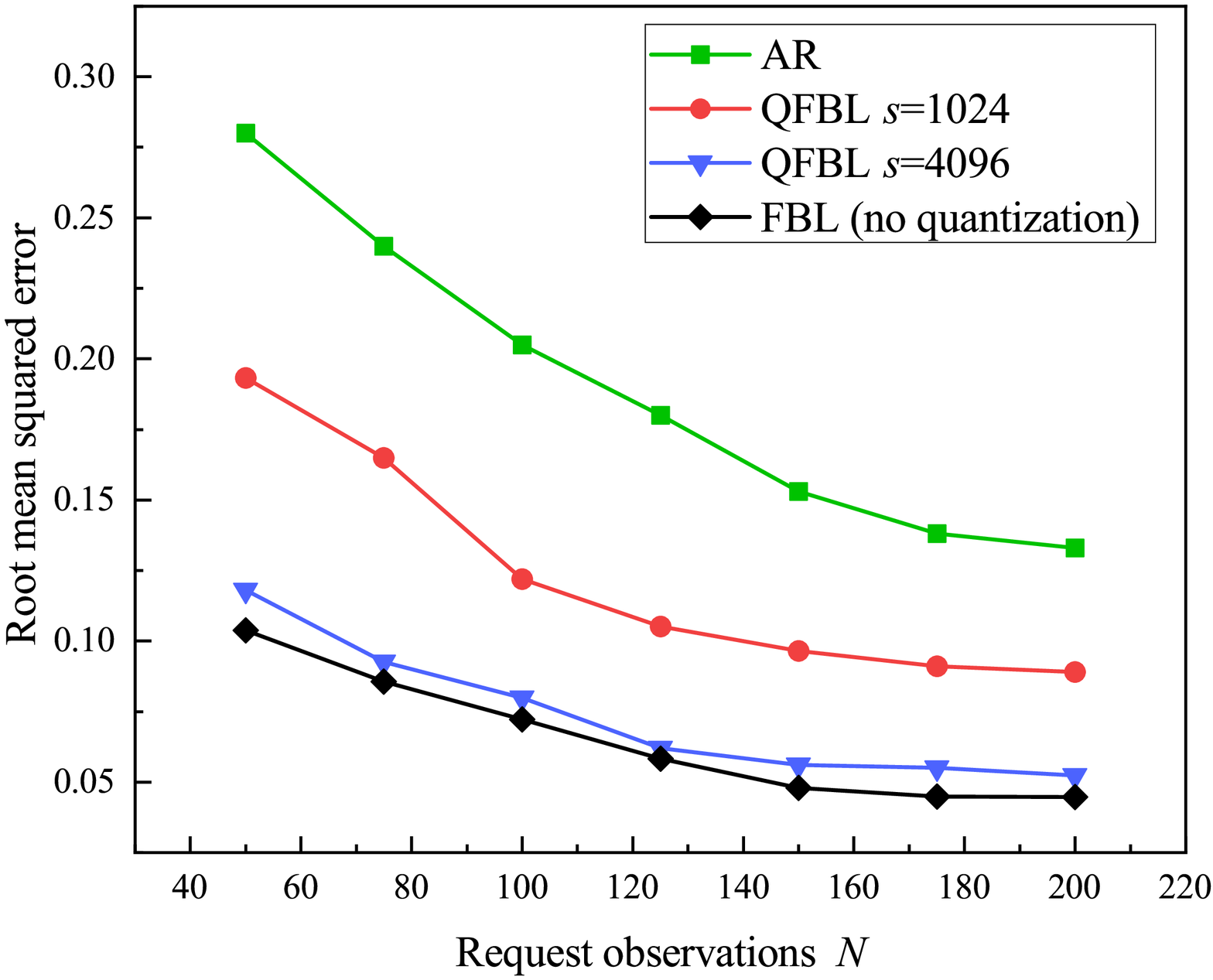}
		\caption{RMSE versus request observations with different $s$.}
		\label{qfbl}
	\end{minipage}
	\begin{minipage}[b!]{0.49\linewidth}
        \vspace{0pt}
		\centering
		\includegraphics[width=0.9\linewidth,height=0.4\linewidth]{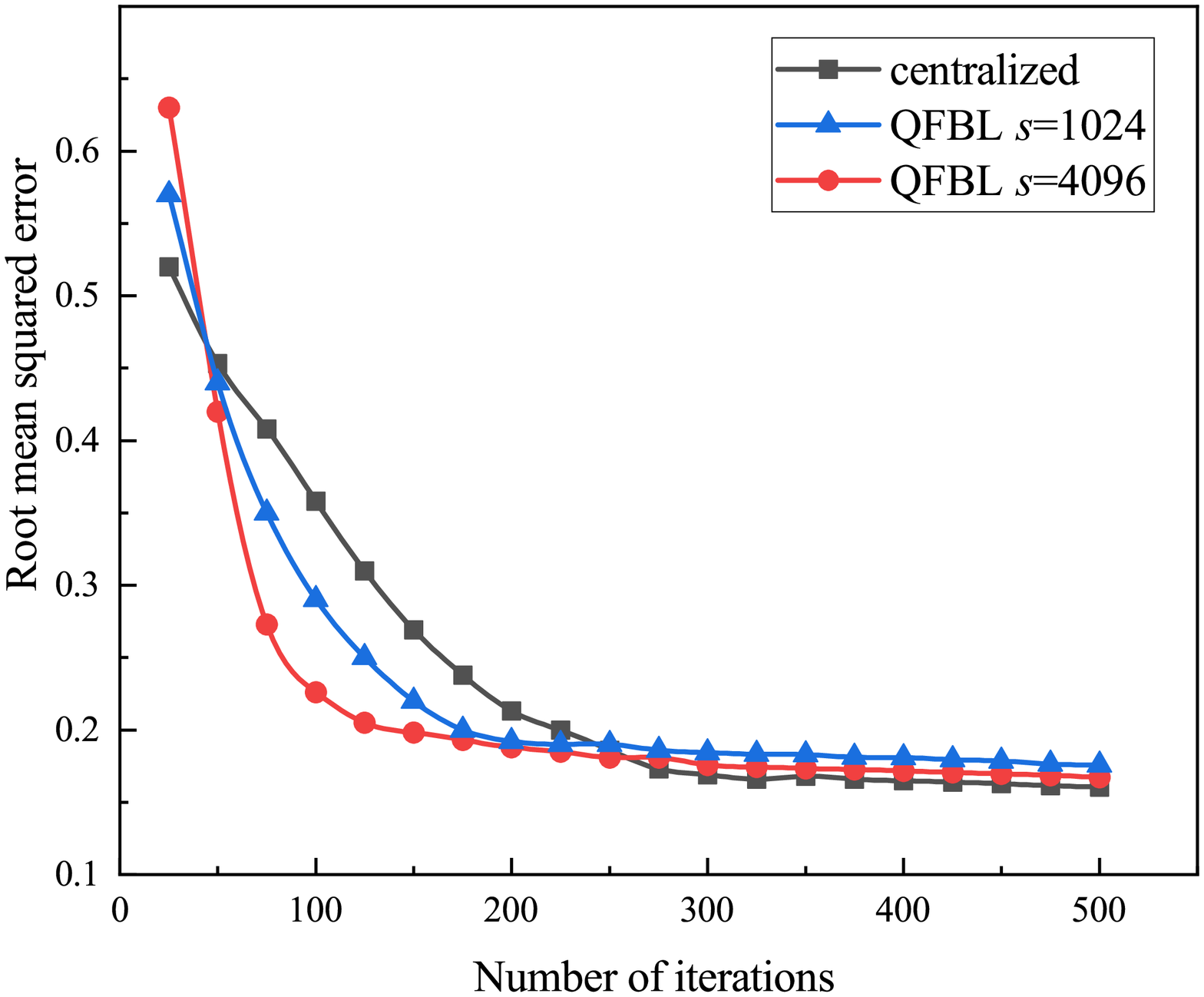}
		\caption{RMSE versus number of iterations.}
		\label{centralized}
	\end{minipage}
\end{figure}

In Fig. \ref{qfbl}, we compare our proposed quantized federated Bayesian learning (QFBL) based predicting policy with the unquantized federated Bayesian learning (FBL) based policy and the auto regressive (AR) process based policy\cite{AR}. It can be seen that similar to what is shown in Fig. \ref{svrggmc}, the RMSE gradually decreases as the observation increases. It can also be seen that when the quantization level $s$ increases, the prediction error is significantly reduced. When $s=1024$, the prediction result is already much better than the AR process based policy. When $s=4096$, the performance of the proposed policy is nearly the same as that of the unquantized case. This is because when $s$ increases, the quantization interval becomes larger, which increases the accuracy of the prediction after quantization.

In Fig. \ref{centralized}, we compare the performance of our proposed policy with centralized learning based policy when the quantization level $s=1024$ and $s=4096$. It can be seen that our proposed policy achieves a high accuracy after a few iterations. This is because FL framework can iterate and then aggregate each existing local model to generate a new global model. The global model can be initialized by the local models, instead of random generation in centralized learning. After reaching convergence, our proposed policy can approach the performance of centralized learning.

\begin{figure}[t!]
	\centering
	\begin{minipage}[t!]{0.49\linewidth}
        \vspace{0pt}
		\centering
		\includegraphics[width=0.9\linewidth,height=0.4\linewidth]{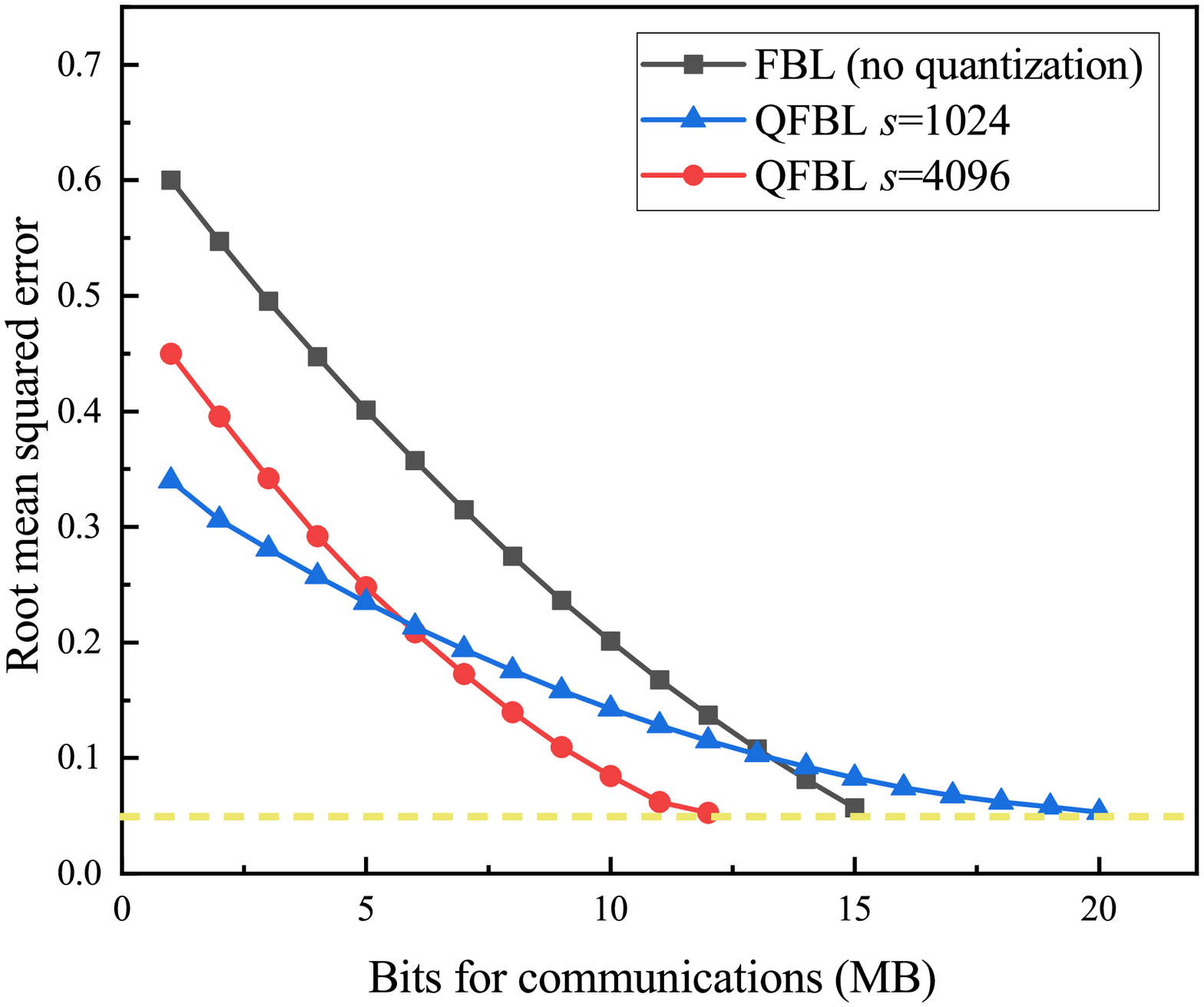}
		\caption{RMSE versus bits for communications.}
		\label{bits}
	\end{minipage}
	\begin{minipage}[t!]{0.49\linewidth}
        \vspace{0pt}
		\centering
		\includegraphics[width=0.9\linewidth,height=0.4\linewidth]{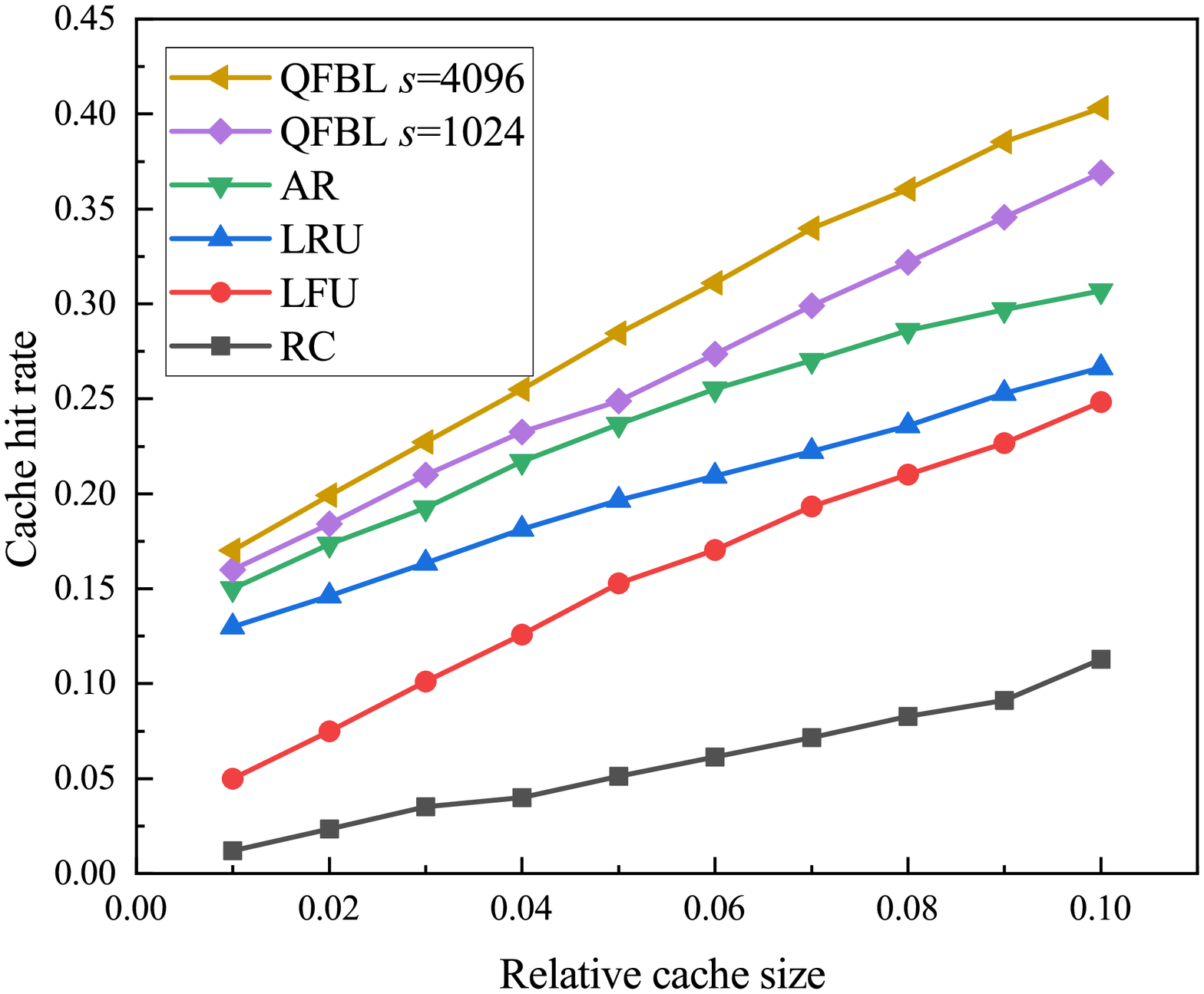}
        \setlength{\abovecaptionskip}{10pt}
		\caption{Cache hit rate versus relative cache size.}
		\label{CHR}
	\end{minipage}
\end{figure}

In Fig. \ref{bits}, we show the effects of quantization on communications. Firstly, we set 0.05 as the satisfying prediction accuracy. We can find that QFBL can get better accuracy with less communications overhead. When $s=4096$, it costs less bits to converge to the predefined accuracy than unquantized FBL. When $s=1024$, the accuracy is higher at the beginning of training. It is because a smaller $s$ allows to perform more iterations using the same number of communications bits, but the convergence is slower than the others due to the larger quantization error brought by a smaller $s$. From the communications perspective, QFBL provides a new method to reduce communications overhead, and it can change the value of $s$ flexibly according to the prediction accuracy required by the system.

In Fig. \ref{CHR}, we show the cache hit rates of our proposed QFBL policy and four traditional policies versus relative cache size. The relative cache size is the ratio of cache size to the size of the content library. The traditional policies include AR process based policy, LRU, LFU and random caching (RC). Assume that the AR process based policy and our proposed policy cache the most popular contents according to the prediction results. It can be observed that the cache hit rates of all policies increase significantly with the relative cache size. It can also be observed that the proposed policy and the AR process based policy have better performance than LRU, LFU and RC due to their utilization of content popularity. Besides, our proposed policy outperforms the AR process based policy by leveraging content features to predict the popularity of newly-added contents.

\section{Conclusions}
In this paper, we have proposed the content popularity prediction policy based on quantized federated Bayesian learning in F-RANs. By using the number of requests and content features, our proposed policy enables prediction of content popularity with lower computational complexity and communications overhead. The reason is that the relationship between content features and popularity is considered in our proposed probabilistic model. At the same time, we have employed FL to leverage the data and computing resources of multiple F-APs to accelerate the convergence. Finally, we have proposed a quantized and encoded federated Bayesian learning sampling method that efficiently trades off between prediction accuracy and communications overhead. We have analyzed the upper bound of the number of bits required by our proposed policy. The quantization level decides the communications overhead of our proposed policy, and a larger quantization level leads to a higher prediction accuracy but more communications overhead. Simulation results have shown that our proposed policy can significantly reduce communications overhead and accelerate the convergence while maintaining the prediction accuracy.

\begin{appendices}
     \section{Proof of Lemma \ref{spar}}
Set $\boldsymbol{u}=\frac{\boldsymbol{v}}{\,\,\left\| \boldsymbol{v} \right\| _2}$ and let $I\left( \boldsymbol{u} \right) $ denote the set of $u_i\leqslant \frac{1}{s}$. Then, we have:
\begin{equation}
1\geqslant \sum_{i\notin I\left( \boldsymbol{u} \right)}^{\,\,}{u_{i}^{2}\geqslant \left( n-\left| I\left( \boldsymbol{u} \right) \right| \right) /s^2}.
\end{equation}
Since $s>0$, it is obvious that $s^2\geqslant n-\left| I\left( \boldsymbol{u} \right) \right|$. Besides, for any $i\in I\left( \boldsymbol{u} \right) $, $Q_i\left( \boldsymbol{v},s \right) $ takes a non-zero value with probability of $u_i$. Therefore, we have:
\begin{equation}
\mathbb{E} \left[ \left\| Q\left( \boldsymbol{v},s \right) \right\| _0 \right] \leqslant d-I\left( \boldsymbol{u} \right) +\sum_{i\in I\left( \boldsymbol{u} \right)u_i}\leqslant s^2+\left\| \boldsymbol{u} \right\| _1\leqslant s^2+\sqrt{d}.
\end{equation}

The unbiasedness can be verified as follows:
\begin{align}
\mathbb{E} \left[ \xi _i\left( \boldsymbol{v},s \right) \right]=\left( l+1-s\frac{\left| v_i \right|}{\left\| \boldsymbol{v} \right\| _2} \right) \cdot \frac{l}{s}+\left( s\frac{\left| v_i \right|}{\left\| \boldsymbol{v} \right\| _2}-l \right) \cdot \frac{l+1}{s}=\frac{\left| v_i \right|}{\left\| \boldsymbol{v} \right\| _2}.
\end{align}

From the definition of $\xi _i\left( \boldsymbol{v},s \right)$, we have:
\begin{align}
\mathbb{E} \left[ \xi _i\left( \boldsymbol{v},s \right) ^2 \right] &=\mathbb{E} \left[ \xi _i\left( \boldsymbol{v},s \right) \right] ^2+\mathbb{E} \left[ \left( \xi _i\left( \boldsymbol{v},s \right) -\mathbb{E} \left[ \xi _i\left( \boldsymbol{v},s \right) \right] \right) ^2 \right] \notag\\
&=\frac{v_{i}^{2}}{\left\| \boldsymbol{v} \right\| _{2}^{2}}+\frac{1}{s^2}q\left( \frac{\left| v_i \right|}{\left\| \boldsymbol{v} \right\| _2},s \right) \left( 1-q\left( \frac{\left| v_i \right|}{\left\| \boldsymbol{v} \right\| _2},s \right) \right) \leqslant \frac{v_{i}^{2}}{\left\| \boldsymbol{v} \right\| _{2}^{2}}+\frac{1}{s^2}q\left( \frac{\left| v_i \right|}{\left\| \boldsymbol{v} \right\| _2},s \right).
\end{align}
Besides, according to $q\left( a,s \right) \leqslant 1$ and $q\left( a,s \right) \leqslant as$, the upper bound of the estimate error can be verified as follows:
\begin{align}
\mathbb{E} \left[ \left\| Q\left( \boldsymbol{v},\boldsymbol{s} \right) \right\| ^2 \right] &=\sum_{i=1}^d{\mathbb{E} \left[ \left\| \boldsymbol{v} \right\| _{2}^{2}\xi _i\left( \frac{\left| v_i \right|}{\left\| \boldsymbol{v} \right\| _2},s \right) ^2 \right]}\leqslant \left\| \boldsymbol{v} \right\| _{2}^{2}\sum_{i=1}^d{\left[ \frac{\left| v_i \right|^2}{\left\| \boldsymbol{v} \right\| _{2}^{2}}+s^2q\left( \frac{\left| v_i \right|}{\left\| \boldsymbol{v} \right\| _2},s \right) \right]}\notag\\
&=\left( 1+\frac{1}{s^2}\sum_{i=1}^d{q\left( \frac{\left| v_i \right|}{\left\| \boldsymbol{v} \right\| _2},s \right)} \right) \left\| \boldsymbol{v} \right\| _{2}^{2}\leqslant \left( 1+\min \left( \frac{d}{s^2},\frac{\left\| \boldsymbol{v} \right\| _1}{s\left\| \boldsymbol{v} \right\| _2} \right) \right) \left\| \boldsymbol{v} \right\| _{2}^{2}\notag\\
&\leqslant \left( 1+\min \left( \frac{d}{s^2},\frac{\sqrt{d}}{s} \right) \right) \left\| \boldsymbol{v} \right\| _{2}^{2}.
\end{align}

This completes the proof.

\section{Proof of Theorem \ref{UpperBound}}
We need two auxiliary lemmas to prove Theorem \ref{UpperBound}.

\newtheorem{lemma4}{Lemma}[section]
\begin{lemma}
For any vector $\boldsymbol{q}\in \mathbb{R} ^d$, $q_i>0$ and $\left\| \boldsymbol{q} \right\| _{p}^{p}\leqslant \rho $, we have:
\begin{equation}
\sum_{i=1}^d{\left| \mathrm{Elias}\left( \boldsymbol{q}_i \right) \right|\leqslant \left( \frac{1+O\left( 1 \right)}{p}\log _2\left( \frac{\rho}{d} \right) +1 \right) d}.
\end{equation}
\label{lem-1}
\end{lemma}
\vspace{-2.5em}
\begin{proof}
For any positive integer $k$, the length of $\mathrm{Elias}\left( k \right) $ is less than $\left( 1+O\left( 1 \right) \right) \log _2k+1$\cite{EliasProperty}. Then, we have:
\begin{align}
\sum_{i=1}^d{\left| \mathrm{Elias}\left( \boldsymbol{q}_i \right) \right|\leqslant} \left( 1+O\left( 1 \right) \right) \sum_{i=1}^d{\left( \log _2q_i \right)}+d\leqslant \frac{\left( 1+O\left( 1 \right) \right)}{p}\sum_{i=1}^d{\left( \log _2\left( q_{i}^{p} \right) \right) +d}.
\end{align}
According to \emph{Jensen inequality}, we have:
\begin{align}
\sum_{i=1}^d{\left| \mathrm{Elias}\left( \boldsymbol{q}_i \right) \right|\leqslant}&\frac{\left( 1+O\left( 1 \right) \right)}{p}d\log _2\left( \frac{1}{d}\sum_{i=1}^d{q_{i}^{p}} \right) +d\notag\\
\leqslant &\frac{\left( 1+O\left( 1 \right) \right)}{p}d\log _2\left( \frac{\rho}{d} \right) +d=\left( \frac{1+O\left( 1 \right)}{p}\log _2\left( \frac{\rho}{d} \right) +1 \right) d.
\end{align}
Therefore, the number of bits required for encoding can be reduced by limiting the number of non-zero elements.
\end{proof}

\newtheorem{lemma5}{Lemma}[section]
\begin{lemma}
For any tuple $\left( A,\boldsymbol{\gamma },\boldsymbol{y} \right) \in \mathcal{B} _s$, the maximum number of bits that $\mathrm{Code}\left( A,\boldsymbol{\gamma },\boldsymbol{y} \right) $ contains can be expressed as follows:
\begin{equation}
F+\left( \left( 1+O\left( 1 \right) \right) \cdot \log _2\left( \frac{d}{\left\| \boldsymbol{y} \right\| _0} \right) +\frac{1+O\left( 1 \right)}{2}\log _2\left( \frac{s^2\left\| \boldsymbol{y} \right\| _{2}^{2}}{\left\| \boldsymbol{y} \right\| _0} \right) +3 \right) \cdot \left\| \boldsymbol{y} \right\| _0.
\end{equation}
\label{lem-1}
\end{lemma}
\vspace{-2.5em}
\begin{proof}
The float number $A$ requires $F$ bits for communications (without Elias coding). $\boldsymbol{\gamma }$ and $\boldsymbol{y}$ can be divided into two parts. First, there is a subsequence $S_1$ defining the non-zero coordinate of $\boldsymbol{y}$. Secondly, there is a subsequence $S_2$ dedicated to representing the sign bit and $c\left( \boldsymbol{y}_i \right) $ for each non-zero coordinate $i$. Although these two subsequences are not consecutive in the string, they clearly divide the remaining bits of the string. The lengths of these two substrings are calculated separately. The information of $\left( A,\boldsymbol{\gamma },\boldsymbol{y} \right) \in \mathcal{B} _s$ can be expressed by $F$, subsequence $S_1$ and $S_2$.

For $S_1$, let $i_1,...,i_{\left\| \boldsymbol{y} \right\| _0}$ denote the positions of the non-zero coordinate in $\boldsymbol{y}$. Then, $S_1$ consists of the encoding of the vector $\boldsymbol{q}^{\left( 1 \right)}=\left[ i_1,i_2-i_1,...,i_{\left\| \boldsymbol{y} \right\| _0}-i_{\left\| \boldsymbol{y} \right\| _0-1} \right]^T$.
Each element of $\boldsymbol{q}^{\left( 1 \right)}$ is encoded with a distance function $c$. According to \emph{Lemma B.1}, since the vector $\boldsymbol{q}^{\left( 1 \right)}$ has length $\left\| \boldsymbol{y} \right\| _0$ and $\left\| \boldsymbol{q}^{\left( 1 \right)} \right\| _1\leqslant d$, we have:
\begin{equation}\label{45}
\left| S_1 \right|\leqslant \left( \left( 1+O\left( 1 \right) \right) \log _2\frac{d}{\left\| \boldsymbol{y} \right\| _0}+1 \right) \left\| \boldsymbol{y} \right\| _0.
\end{equation}

For $S_2$, each non-zero coordinate is required to pass the sign and the encoding result of $sy_i$, so the length of $S_2$ can be determined by the two parts. According to \emph{Lemma B.1}, we have:
\begin{align}\label{46}
\left| S_2 \right|=\sum_{j=1}^{\left\| \boldsymbol{y} \right\| _0}{\left( 1+\left| \mathrm{Elias}\left( sy_j \right) \right| \right)}\leqslant \left\| \boldsymbol{y} \right\| _0+\left( \frac{1+O\left( 1 \right)}{2}\log _2\frac{s^2\left\| \boldsymbol{y} \right\| _{2}^{2}}{\left\| \boldsymbol{y} \right\| _0}+1 \right) \left\| \boldsymbol{y} \right\| _0.
\end{align}
According to Eq. (\ref{45}) and Eq. (\ref{46}), \emph{Lemma B.2} can then be proved.
\end{proof}

Based on \emph{Lemma B.2} and the sparsity of $Q\left( \boldsymbol{v},s \right) $, we have:
\begin{align}
\label{aa}
\mathbb{E} \left[ \left| \mathrm{Code}\left( Q\left( \boldsymbol{v},s \right) \right) \right| \right] &\leqslant F+\left( 1+O\left( 1 \right) \right) \mathbb{E} \left[ \left\| \boldsymbol{\varepsilon } \right\| _0\log _2\left( \frac{d}{\left\| \boldsymbol{\varepsilon } \right\| _0} \right) \right] \notag\\
&+\frac{1+O\left( 1 \right)}{2}\mathbb{E} \left[ \left\| \boldsymbol{\varepsilon } \right\| _0\log _2\left( \frac{s^2\left\| \boldsymbol{\varepsilon } \right\| _{2}^{2}}{\left\| \boldsymbol{\varepsilon } \right\| _0} \right) \right] +3\mathbb{E} \left[ \left\| \boldsymbol{\varepsilon } \right\| _0 \right] \notag\\
&\leqslant F+\left( 1+O\left( 1 \right) \right) \mathbb{E} \left[ \left\| \boldsymbol{\varepsilon } \right\| _0\log _2\left( \frac{d}{\left\| \boldsymbol{\varepsilon } \right\| _0} \right) \right] \notag\\
&+\frac{1+O\left( 1 \right)}{2}\mathbb{E} \left[ \left\| \boldsymbol{\varepsilon } \right\| _0\log _2\left( \frac{2\left( s^2+d \right)}{\left\| \boldsymbol{\varepsilon } \right\| _0} \right) \right] +3\left( s^2+\sqrt{d} \right).
\end{align}

Clearly, $f\left( x \right) =x\log _2\left( \frac{C}{x} \right) $ is a concave function at $C>0$, which is an increasing function until $x=\frac{C}{2}$ and then decreases gradually. According to \emph{Jensen's inequality}, the sparsity of $Q\left( \boldsymbol{v},s \right) $ and the assumption $s^2+\sqrt{d}\leqslant \frac{d}{2}$, we have:
\begin{align}
\label{47}\mathbb{E} \left[ \left\| \boldsymbol{\varepsilon } \right\| _0\log _2\left( \frac{d}{\left\| \boldsymbol{\varepsilon } \right\| _0} \right) \right] &\leqslant \left( s^2+\sqrt{d} \right) \log _2\left( \frac{d}{s^2+\sqrt{d}} \right),\\
\label{48}\mathbb{E} \left[ \left\| \boldsymbol{\varepsilon } \right\| _0\log _2\left( \frac{2\left( s^2+d \right)}{\left\| \boldsymbol{\varepsilon } \right\| _0} \right) \right] &\leqslant \left( s^2+\sqrt{d} \right)\cdot
\log _2\left( \frac{2\left( s^2+d \right)}{s^2+\sqrt{d}} \right).
\end{align}
Then, \emph{Theorem \ref{UpperBound}} can be proved by combining Eq. (\ref{aa}), Eq. (\ref{47}) and Eq. (\ref{48}).

This completes the proof.

\end{appendices}

\balance
\bibliographystyle{IEEEtran}
\bibliography{myref}
\end{document}